%% file: main.tex
\documentclass[10pt,twocolumn,letterpaper]{article}

\usepackage[pagenumbers]{cvpr} 
\usepackage[most]{tcolorbox}
\input{preamble}

\definecolor{cvprblue}{rgb}{0.21,0.49,0.74}
\usepackage[pagebackref,breaklinks,colorlinks,allcolors=cvprblue]{hyperref}

\def\paperID{XXXX} 
\def\confName{CVPR}
\def\confYear{2026}

\title{VLM-3R: Vision-Language Models \\ Augmented with Instruction-Aligned 3D Reconstruction}

\author{
Zhiwen Fan$^{1*\dagger}$\hspace{0.4em}
Jian Zhang$^{2*}$\hspace{0.4em}
Renjie Li$^3$\hspace{0.4em}
Junge Zhang$^4$\hspace{0.4em}
Runjin Chen$^1$\hspace{0.4em}
Hezhen Hu$^1$\hspace{0.4em}
Kevin Wang$^1$\\
Peihao Wang$^1$\hspace{0.4em}
Huaizhi Qu$^5$\hspace{0.4em}
Shijie Zhou$^7$\hspace{0.4em}
Dilin Wang$^6$\hspace{0.4em}
Zhicheng Yan$^6$\hspace{0.4em}
Hongyu Xu$^6$\\
Justin Theiss$^6$\hspace{0.4em}
Tianlong Chen$^5$\hspace{0.4em}
Jiachen Li$^4$\hspace{0.4em}
Zhengzhong Tu$^3$\hspace{0.4em}
Zhangyang Wang$^{1\dagger}$\hspace{0.4em}
Rakesh Ranjan$^{6\dagger}$\\[0.3em]
\normalsize
$^1$UT Austin \hspace{0.6em}
$^2$XMU \hspace{0.6em}
$^3$TAMU \hspace{0.6em}
$^4$UCR \hspace{0.6em}
$^5$UNC \hspace{0.6em}
$^6$Meta \hspace{0.6em}
$^7$UCLA\\[0.3em]
\url{https://vlm-3r.github.io/}
}

\begin{document}

\twocolumn[{%
\renewcommand\twocolumn[1][]{#1}%
\maketitle

\vspace{-12mm}
\begin{center}
    \includegraphics[width=\textwidth]{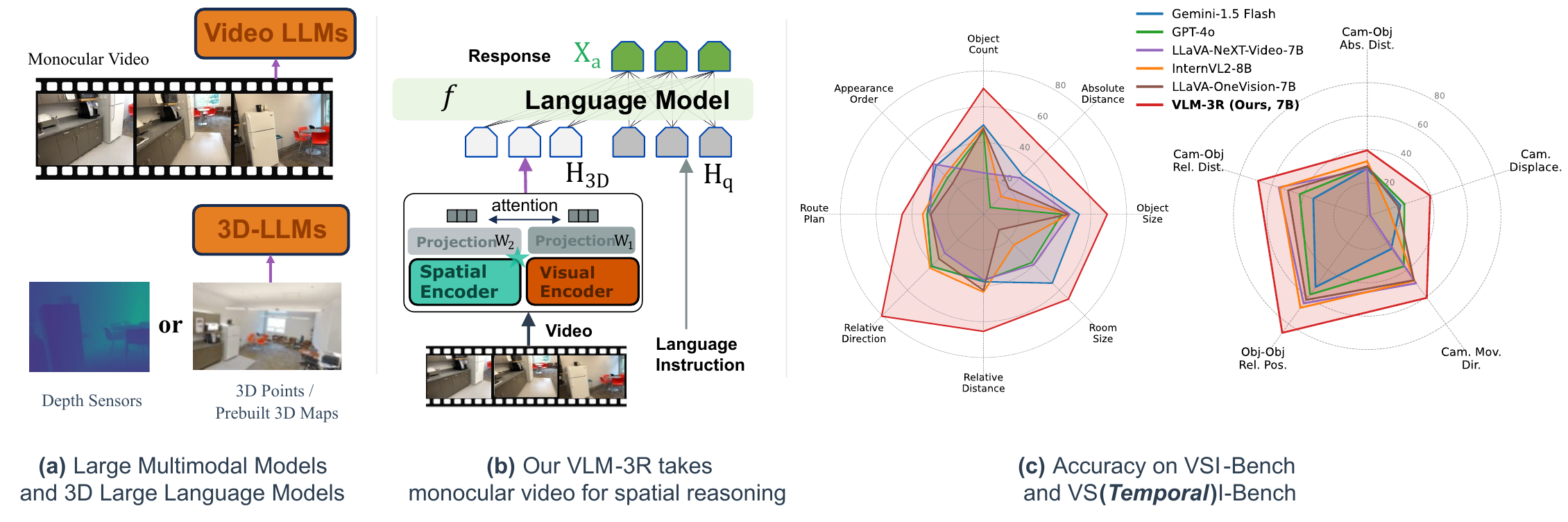}
    \vspace{-4mm}
    \captionof{figure}{
      \textbf{VLM-3R: 3D Spatial-Temporal Reasoning from Monocular Video.}
Unlike prior video LMMs and 3D-LLMs (a) that depend on explicit 3D inputs such as depth sensors or pre-built 3D maps, VLM-3R (b) uses an end-to-end architecture with metric-scale spatial encoders that fuse scene and camera tokens to recover implicit 3D structure directly from monocular video.
This design targets spatial assistance from raw camera streams and supports detailed reasoning about spatial context, instance layout, and temporal dynamics, leading to consistently stronger performance on VSI-Bench and the proposed VS\textit{Temporal}I-Bench (c).
    }
    \label{fig:teaser}
\end{center}
}]

\begingroup
\renewcommand{\thefootnote}{\fnsymbol{footnote}}
\footnotetext[1]{Equal contribution. $\dagger$ Corresponding Author.}
\endgroup

\input{sec/0_abstract}    
\input{sec/1_intro}
\input{sec/2_related_work}

\input{sec/3_data}

\input{sec/4_method}

\input{sec/5_experiment}
\input{sec/6_conclusion}

\clearpage
{
    \small
    \bibliographystyle{ieeenat_fullname}
    \bibliography{main}
}

\input{sec/X_suppl}

\end{document}

%% file: preamble.tex
\usepackage{multirow}
\usepackage{bbm}
\usepackage{graphicx}   
\usepackage{xcolor}     
\usepackage{colortbl}   
\usepackage{array}      
\usepackage{booktabs}   
\usepackage[T1]{fontenc}    
\usepackage{makecell} 

\usepackage{caption}
\usepackage[table]{xcolor} 
\usepackage{pifont}        
\newcommand{\cmark}{\ding{51}} 
\newcommand{\xmark}{\ding{55}} 

\definecolor{navyblue}{HTML}{0071BC}
\definecolor{hotpink}{HTML}{FF0080}
\definecolor{oai-white}{HTML}{FFFFFF}
\definecolor{oai-black}{HTML}{000000}
\definecolor{oai-red}{HTML}{FF4500}
\definecolor{oai-green}{HTML}{51DA4C}
\definecolor{oai-blue}{HTML}{0000FF}
\definecolor{oai-yellow}{HTML}{FFF639}
\definecolor{oai-magenta}{HTML}{FF45FF}
\definecolor{oai-cyan}{HTML}{00FFFF}
\definecolor{oai-orange}{HTML}{FE7600}
\definecolor{oai-violet}{HTML}{8A2BE2}
\definecolor{oai-brown}{HTML}{A0522D}
\definecolor{oai-green-050}{HTML}{F4FFF4}
\definecolor{oai-green-100}{HTML}{E9FFE8}
\definecolor{oai-green-200}{HTML}{D9FFD8}
\definecolor{oai-green-300}{HTML}{C9FFC7}
\definecolor{oai-green-400}{HTML}{A6FFA3}
\definecolor{oai-green-500}{HTML}{7CF178}
\definecolor{oai-green-600}{HTML}{51DA4C}
\definecolor{oai-green-700}{HTML}{3FA93B}
\definecolor{oai-green-800}{HTML}{2D712A}
\definecolor{oai-green-900}{HTML}{193718}
\definecolor{oai-gray-000}{HTML}{FFFFFF}
\definecolor{oai-gray-100}{HTML}{FAFAFA}
\definecolor{oai-gray-200}{HTML}{F5F5F5}
\definecolor{oai-gray-300}{HTML}{E5E5E5}
\definecolor{oai-gray-400}{HTML}{FFB7A4}
\definecolor{oai-gray-500}{HTML}{CDCDCD}
\definecolor{oai-gray-600}{HTML}{A8A8A8}
\definecolor{oai-gray-700}{HTML}{747474}
\definecolor{oai-gray-800}{HTML}{393939}
\definecolor{oai-gray-900}{HTML}{000000}



%% file: sec/0_abstract.tex
\begin{abstract}
The rapid advancement of Large Multimodal Models (LMMs) for 2D images and videos has sparked interest in extending these models to 3D scenes, with the goal of human-like visual-spatial intelligence.
However, achieving deep spatial understanding comparable to human capabilities remains challenging for both model design and data acquisition.
Existing methods often rely on external depth sensors for geometry capture or off-the-shelf algorithms for pre-constructing 3D maps, which limits their scalability.
In this work, we introduce VLM-3R, a framework for Vision-Language Models (\textbf{\underline{VLM}}s) that couples \textbf{\underline{3}}D \textbf{\underline{R}}econstructive instruction tuning with scalable training data curation and a new benchmark for temporal reasoning.
Specifically, VLM-3R processes monocular video frames with a geometry encoder that derives implicit 3D tokens representing scene context (spatial tokens) and camera motion (view tokens).
In parallel, we build a scalable data creation pipeline with over 200K 3D reconstructive instruction-tuning question-answer (QA) pairs.
To evaluate temporal reasoning, we further introduce the Vision-Spatial-Temporal Intelligence benchmark (VSTI-Bench), which contains over 138.6K QA pairs across five distinct tasks focused on evolving spatial relationships.
Extensive experiments show that VLM-3R supports robust visual-spatial reasoning and improves the understanding of temporal 3D context changes, enabling monocular 3D spatial assistance and embodied reasoning.

\end{abstract}
\vspace{-4mm}

%% file: sec/1_intro.tex
\section{Introduction}

Humans acquire visual-spatial intelligence through visually guided interaction with the physical world~\cite{adolph2000specificity}, forming and updating internal maps and recalling experiences about spatial layouts, object sizes, possible actions, and the timing of events.
By contrast, Large Language Models (LLMs~\cite{naveed2024comprehensiveoverviewlargelanguage, wei2022emergent, beguvs2023large, zhang-etal-2024-unveiling-linguistic, Kassner2023LanguageMW,radford2018improving,radford2019language,brown2020language,touvron2023llama,touvron2023llama2,bai2023qwen,team2023gemini}) have made strong progress on complex text reasoning, and Vision-Language Models (VLMs) and Large Multimodal Models (LMMs~\cite{hurst2024gpto,team2024gemini,alayrac2022flamingo,li2023blip2,liu2024visual,bai2023qwenvl,chen2024internvl,wang2024survey}) show solid ability in open-ended dialogue, single-image understanding, and practical tasks such as web agents. Despite these gains, current VLMs and LMMs still fall short when interacting with the physical world, where they struggle with core spatial skills that require geometric understanding, even for simple tasks like distance estimation.

Existing efforts to advance LLMs and VLMs for spatial tasks~\cite{chen2024spatialvlm,cheng2024spatialrgpt,hong20233d} often rely on specialized depth sensors~\cite{zhu2024llava,zheng2024video} for 3D data during fine-tuning and inference, which constrains scalability to sensor-equipped environments and restricts the use of abundant monocular video data. Other works~\cite{mao2025spatiallm} employ off-the-shelf reconstruction or SLAM algorithms~\cite{murai2025mast3r,hong20233d,hong2024multiply} to pre-construct explicit 3D maps, typically point clouds, which are then aligned with or fed as input to language models.
This also introduces persistent challenges: the reliance on pre-constructed geometry makes these multi-stage pipelines slow to adapt and reason in new spaces for a generalist model, and they are brittle when traditional reconstruction ignores real-world scene scale~\cite{schonberger2016structure}, which can irreversibly degrade the VLM's spatial comprehension.

In this paper, we propose VLM-3R, which equips VLMs with 3D reconstructive instruction tuning from monocular video. It addresses the central question of how to build an end-to-end, spatially aware VLM system that interprets scene geometry, camera movement, and spatial relations from image sequences without requiring intermediate runtime reconstruction, while also integrating common-sense knowledge from language models for spatial assistance and embodied reasoning.
Specifically, VLM-3R first proposes a framework that unifies image-level semantics from a vision encoder with rich spatial information from a metric-scale multiview stereo model~\cite{wang2025continuous} into a geometry encoder. We extract implicit 3D tokens representing scene context (spatial tokens) and camera motion (view tokens) from this geometry encoder. These physical scale and camera movement priors are then integrated with the original visual encoding via \textit{Spatial-Visual-View Fusion}, aligning them with language representations for joint spatio-linguistic understanding within the VLM. This enables the model to interpret spatial context directly from raw video and to disentangle changes in camera-object relative distance. Complementing the architecture design, we also develop a scalable 3D video data curation pipeline to support 3D reconstructive instruction tuning with diverse spatial tasks.

To further examine the capability of existing VLMs for \textbf{spatio-temporal} understanding from video, we introduce the Vision-Spatial-\textit{Temporal} Intelligence benchmark (VSTI-Bench), which comprises five temporal tasks defined on \emph{static} 3D scenes. Since the scenes are static, all apparent motion arises from camera movement. Accordingly, the benchmark evaluates whether models can reason about temporal changes in spatial relations induced by camera motion, including camera displacement over time, changes in camera-object relative distance and direction, and frame-dependent object-object relative position from the camera's perspective. Experiments show that VLM-3R not only reasons about static spatial context, but also effectively captures spatio-temporal changes in 3D environments from monocular video input, as evidenced by its performance across benchmarks in Figure~\ref{fig:teaser}(c).
To summarize, our contributions are:

\begin{itemize}
\item We introduce VLM-3R, a method that significantly enhances the visual-based spatial intelligence of existing vision-language frameworks, demonstrating strong spatial understanding and reasoning capabilities from monocular video without requiring any depth sensors or pre-built 3D maps as input.

\item We design Spatial-Visual-View Fusion, which injects scene geometry and camera motion from a feed-forward geometry encoder into image-level semantics, supporting both spatial and temporal reasoning in the model.

\item We build a scalable data curation pipeline that produces 3D-aware instruction data with labels for spatial relation discovery, totaling over 200K spatial reconstructive QA pairs with 4{,}225 route-planning instances. We further study VLM understanding of temporal changes and introduce a new benchmark with 138.6K temporal QA pairs.

\item Extensive experiments on VSI-Bench, VSTI-Bench, and OST-Bench show state-of-the-art performance on both visual spatial and temporal understanding tasks, approaching baselines that use ground-truth depth while maintaining strong generalization across scenes and tasks.
\end{itemize}

%% file: sec/2_related_work.tex
\section{Related Work}
\paragraph{Large Multimodal Models.}
Large Multimodal Models (LMMs) aim to unify vision, language, and other modalities into a single model. Early models like CLIP~\cite{radford2021clip} and ALIGN~\cite{jia2021align} learned joint image-text representations via contrastive pretraining, enabling strong zero-shot performance. Later models such as Flamingo~\cite{alayrac2022flamingo} and BLIP-2~\cite{li2023blip} improved efficiency by decoupling vision and language modules, enhancing cross-modal reasoning. Recent works have expanded LMMs to broader applications, including embodied agents (PaLM-E~\cite{driess2023palm}), grounding~\cite{peng2023kosmos}, and general-purpose task solving~\cite{lu2022unified,wang2022ofa}.
Several recent efforts~\cite{zhou2024feature,zhou2025feature4x,deng20253d,zhu2024llava,hong20233d,zheng2024video} have extended LMMs with explicit or implicit 3D representations for spatial understanding. LLaVA-3D~\cite{zhu2024llava} and ROSS3D~\cite{wang2025ross3d} incorporate multi-view images and depth supervision to improve RGB-D spatial QA, while SpatialRGPT~\cite{cheng2024spatialrgpt} enhances spatial comprehension by injecting regional features augmented with monocular depth. Building upon VLM architectures, Spatial-MLLM~\cite{chen2024spatialvlm}, VG-LLM~\cite{deng20253d}, and SpatialStack~\cite{zhang2026spatialstack} introduce a multi-view geometry transformer VGGT~\cite{wang2025vggt} and fuse its geometry-aware tokens with vision and language features to strengthen 3D reasoning.
However, VGGT predicts only normalized depth and lacks global metric scale, which leads to scale-ambiguous geometric representations that hinder accurate spatial estimation such as metric distance and object size reasoning. In addition, many geometry-augmented VLMs rely on RGB-D sensors, precomputed geometric maps, or auxiliary depth pipelines, and therefore have limited applicability in real-world monocular video scenarios. To address these limitations, we leverage CUT3R~\cite{wang2025continuous}, a multi-view geometry model that reconstructs globally aligned metric-scale 3D structures directly from monocular video, enabling stronger absolute-scale perception in realistic environments.

\vspace{-4mm}
\paragraph{Spatial Reasoning for Visual-Spatial Intelligence.}
Spatial reasoning is crucial to human cognition, enabling us to navigate complex environments~\cite{gardner2011frames,mcnamara2002locations,waller2013handbook}. Achieving this in machines, especially from monocular RGB video, remains challenging. Benchmarks like VSI-Bench~\cite{yang2024thinking} assess LMMs' ability to understand spatial relations in real-world videos, with eight tasks spanning configurational, measurement estimation, and spatiotemporal categories, testing the model's grasp of intuitive spatial configurations, while recent works such as VLM4D~\cite{zhou2025vlm4d}, Thinking in Dynamics~\cite{huang2026thinkingdynamicsmultimodallarge}, and DynamicVerse~\cite{wen2025dynamicverse} extend these evaluations to dynamic scenes.
Despite progress in 2D visual perception, state-of-the-art models like Gemini 1.5~\cite{team2024gemini}, GPT-4o~\cite{hurst2024gpt}, and open-source models such as InternVL~\cite{chen2024far}, ViLA~\cite{lin2024vila}, LongViLA~\cite{chen2024longvila}, LongVA~\cite{zhang2024long}, and LLaVA-OneVision~\cite{li2024llava} still show substantial gaps in spatial tasks like localization, layout inference, and memory recall. These models' shortcomings in spatial reasoning stem from the lack of structured spatial representations and multi-view encoding, which are crucial for human-like perception. Recent dual-encoder spatial VLMs~\cite{chen2024spatialvlm,deng20253d} partially address this by incorporating geometric encoders, but their normalized depth representation lacks metric scale, making accurate measurement-dependent reasoning difficult.
This gap further highlights the importance of spatial schemas and hierarchical scene encoding in human cognition~\cite{hegarty2005individual,meneghetti2022individual,waller2013handbook}. We propose an end-to-end approach that extracts rich 3D spatial information from monocular RGB video and seamlessly integrates it into existing LMMs, enabling robust spatial reasoning without modular pipelines or additional depth sensors.


\vspace{-4mm}
\paragraph{3D Reconstruction from Images.}
Classical pipelines rely on modular SfM+MVS stages (feature extraction, matching, geometric verification)~\cite{hartley2003multiple,schonberger2016structure}, which are highly accurate but notoriously slow to optimize in practice. Learnable MVS methods, often inspired by MVSNet~\cite{yao2018mvsnet, gu2020cascade}, instead predict depth maps from multi-view images given known cameras. Recent transformer-based approaches such as DUSt3R~\cite{wang2024dust3r} and MASt3R~\cite{leroy2024grounding} further relax camera requirements by directly regressing pixel-aligned 3D point maps, and have been rapidly extended to multi-view reconstruction~\cite{tang2024mv,wang20243d,wang2025vggt,yang2025fast3r}, semantic understanding~\cite{fan2024large}, and dynamic scenes~\cite{zhang2024monst3r,wang2025continuous}.

%% file: sec/3_data.tex
\section{Scalable Spatial and Temporal Data}

\begin{figure}[!t]
  \centering
  \includegraphics[width=0.99\linewidth]{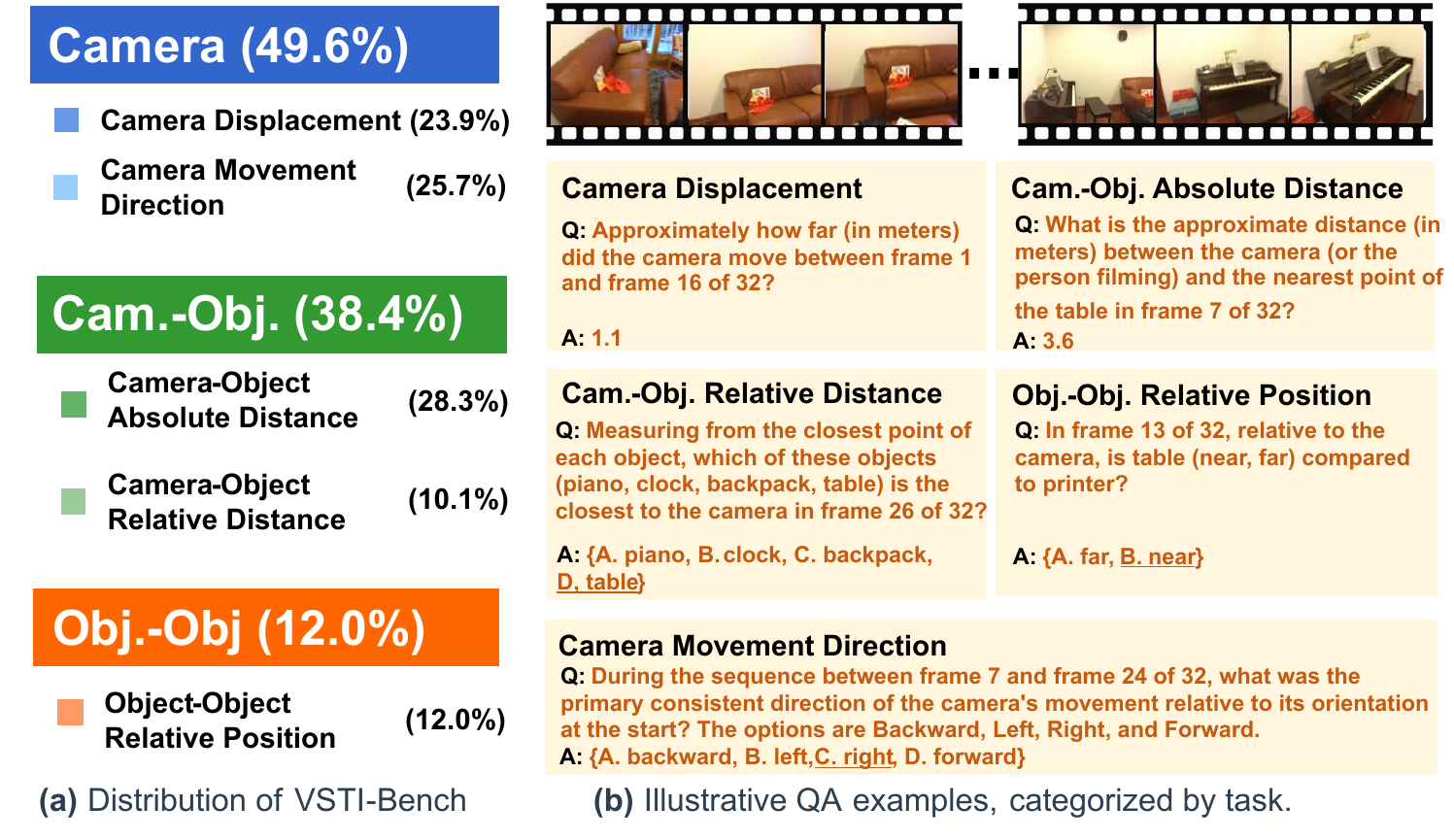} 
  \vspace{-3mm}
\caption{\textbf{VS\textit{Temporal}I-Bench Overview.} It contains 138.6K QA samples across multiple spatio-temporal question types. (a) Statistical distribution of QA pairs by primary categories (inner ring, detailed in the legend) and their sub-categories (outer ring). (b) Example QA pairs illustrating different task types within a benchmark scene.}
\vspace{-5mm}
\label{fig:vsti-stats}
\end{figure}

\subsection{Overview}
Earlier work VSI-Bench~\cite{yang2024thinking} introduced thousands of visual-spatial question-answer (QA) pairs through manual annotation and semi-automated tools, but these methods are difficult to scale. 
We build a scalable pipeline for curating diverse, large-scale data to improve spatial QA capabilities, using an automated process on existing 3D datasets containing video or multi-view images, aided by simulators to create complex route-planning scenarios. 
Crucially, to prevent any potential data leakage between our training and evaluation sets, our pipeline strictly adheres to the standard train/test splits provided by these source datasets.

We also study the spatio-temporal understanding of existing VLMs by creating a new benchmark, VSTI-Bench, with 138.6K data samples, including both training and testing splits, that focuses on reasoning about spatial configurations evolving over time (Figure~\ref{fig:vsti-stats}). 




\subsection{Multimodal Instruction Generation} \label{sec:data-vsi}
VSI-Bench~\cite{yang2024thinking} provides about 5{,}000 question-answer pairs from real-world videos, but this scale is insufficient to advance robust spatial reasoning in VLMs. Many current models, including large proprietary systems, still struggle with basic spatial tasks such as inter-object distance estimation, indicating limited spatial supervision and encoding capacity. We address these gaps with a scalable and highly automated data pipeline that generates over \textbf{200K} diverse spatial reasoning QA pairs from monocular video and \textbf{4{,}225} simulator-based embodied route-planning instances to significantly strengthen spatial intelligence in LMMs.

\vspace{-4mm}
\paragraph{General Spatial Question-Answer Generation}
We consolidate diverse data sources into a unified meta-information structure. For open-source 3D datasets that provide 3D geometry, semantics, and instance-level meta-information, such as ScanNet~\cite{dai2017scannet}, ScanNet++~\cite{yeshwanth2023scannet++}, and ARKitScenes~\cite{baruch2021arkitscenes}, we derive QA data that target seven of the eight core task types in VSI-Bench. The foundation for our QA generation is a detailed spatio-temporal scene graph (see more details in the supplementary material). In this graph, each frame is treated as a temporal node, and each object instance is a distinct node with attributes such as global and local coordinates and detailed semantic information. This scene graph enables us to determine the precise global and local status of each instance, including its relation to the camera, at every time step. Leveraging these rich temporal relationships, we automatically generate question-answer pairs for seven key tasks.


\vspace{-4mm}
\paragraph{Route Planning Data Generation}
The Route Planning task involves producing high-level navigation descriptions within a given environment. Although VSI-Bench~\cite{yang2024thinking} includes 194 human-annotated instances, this scale is insufficient for robust training. To enable effective spatial route planning, we employ the Habitat simulator~\cite{habitat} to generate accurate and plausible navigation routes at scale.

In this simulation setup, an agent starts at a specified location and receives a human-generated navigation instruction describing a path to a target goal. The agent follows this instruction by executing a sequence of discrete actions (e.g., turning, moving forward) to reach the goal, concluding with a \textit{stop} action upon arrival or completion. During navigation, we sample thousands of navigable paths in 3D scenes and generate corresponding textual descriptions based on the agent's position, orientation (including visible objects), and traversed path segments. Object descriptions use accurate 3D ground-truth labels, selecting annotations relevant to the agent's current viewpoint and location. Finally, these textual descriptions are used to construct 4{,}225 QA pairs that conform to the VSI-Bench route planning format.

\subsection{New Spatio-Temporal Reasoning Benchmark}

To move beyond static understanding of 3D environments, we introduce the Vision-Spatial-Temporal Intelligence benchmark (\textbf{VSTI-Bench}). This benchmark is designed to test whether AI models can not only answer global questions from input video but also reason about objects, cameras, and their relationships as they evolve over time. Its primary goal is to evaluate and improve the spatio-temporal reasoning abilities of egocentric video-based Large Multimodal Models (LMMs). Figure~\ref{fig:vsti-stats} depicts the distribution of tasks together with representative examples.

\vspace{-4mm}

\paragraph{Task Definition: Temporal Reasoning}
While VLMs can process static scenes effectively, evaluating temporal reasoning from monocular video is essential for advancing spatio-temporal understanding. Our VS\textit{Temporal}I-Bench is designed to probe an LMM's ability to perceive and reason about camera motion, camera–object interactions, and evolving spatial configurations. As shown in Figure~\ref{fig:vsti-stats}(a), it contains approximately 138,600 QA pairs across three primary categories: Camera Dynamics (49.6\%), Camera–Object Interactions (38.4\%), and Object Relative Position (12.0\%). Additional details for each category are provided in the supplementary materials.

\begin{figure*}[!t]
\centering
\includegraphics[width=0.85\linewidth]{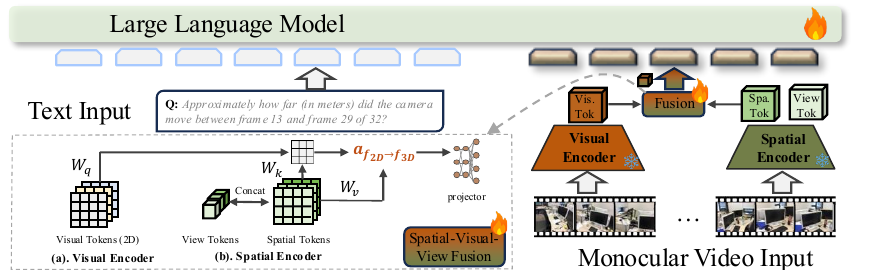}
\vspace{-3mm}
\caption{\textbf{Network Architecture.} VLM-3R takes monocular video and language instructions as input. A vision encoder and metric-scale geometry encoder extract frame-level appearance, camera-view pose, and globally aligned 3D structure. Spatial-Visual-View Fusion applies 2D-3D attention and a layer projector to inject spatial tokens and view tokens into the VLM. At inference time, the model uses these fused representations to support reliable spatial and temporal reasoning from monocular video, without requiring depth sensors or pre-built 3D maps.}
\label{fig:arc}
\vspace{-3mm}
\end{figure*}

\vspace{-4mm}
\paragraph{Temporal Question-Answer Generation}
The question-answer pairs for VS\textit{Temporal}I-Bench originate from diverse 3D datasets (ScanNet~\cite{dai2017scannet}, ScanNet++~\cite{yeshwanth2023scannet++}, and ARKitScenes~\cite{baruch2021arkitscenes}), which feature videos captured by hand-held monocular cameras, often accompanied by ground-truth depth maps and instance annotations from varied environments. To generate these QA pairs, we construct a detailed spatio-temporal scene graph as described in the previous section. This graph explicitly models each object instance with its geometric attributes (e.g., position, size, rotation) and semantic labels, together with precise camera viewpoint information (pose) at every relevant time step, thereby capturing the global and local status of instances and their evolving associations with the camera. By leveraging this rich scene graph, we can accurately compute various temporal changes (such as net camera displacement) that define our tasks. This structured understanding then enables the automated formulation of diverse question-answer pairs that probe camera dynamics, object states, and complex camera-object interactions over time, as exemplified in Figure~\ref{fig:vsti-stats}(b).


\vspace{-4mm}
\paragraph{Metrics}
For Multiple-Choice Answer (MCA) tasks we use standard Accuracy (ACC)~\cite{fu2024video,hendrycks2020measuring,yue2024mmmu} via exact or fuzzy matching. For Numerical Answer (NA) tasks we adopt \emph{Mean Relative Accuracy} (MRA)~\cite{yang2024thinking}, defined as
$\mathrm{MRA} = \frac{1}{10}\sum_{\theta\in{0.5,0.55,\dots,0.95}}\mathbbm{1}\bigl(|\hat y - y|/y < 1-\theta\bigr)$,
which captures prediction proximity across multiple tolerance levels.

%% file: sec/4_method.tex
\section{VLM-3R Architecture}

\paragraph{Overview}
Figure~\ref{fig:arc} illustrates the architecture of VLM-3R. During both training and inference, the primary inputs are a monocular video, represented as a sequence of frames $\{I_t\}_{t=1}^N$ (where each frame $I_t \in \mathbb{R}^{H \times W \times 3}$), and accompanying language instructions. The model extracts visual, geometric, and camera pose tokens from this video via its end-to-end architecture; these tokens are subsequently aligned with language representations using instruction tuning, facilitated by our curated data and a few learnable layers.

\subsection{Geometry-Aware Vision-Language Model}
VLM-3R retains the generality of VLM architectures such as \cite{zhang2024llavanextvideo} while not requiring additional depth or 3D maps as input. It integrates geometric and camera-view encodings via a geometry encoder and visual features via a visual encoder from the input video, then fuses these signals with language representations..

\vspace{-3mm}
\paragraph{3D Reconstructive Tokenization.}\label{sec:3dtoken}
VLM-3R adopts the pre-trained CUT3R model~\cite{wang2025continuous} as a core component for 3D reconstructive tokenization. CUT3R processes monocular video frame-by-frame: each incoming image $I_t$ is first passed through an image encoder $f_{\text{enc}}$ (e.g., a Vision Transformer) to extract feature tokens $F_t$. Subsequently, these tokens $F_t$, along with a learnable pose query token $z$ and the previous recurrent state $s_{t-1}$, are processed by a transformer decoder $f_{\text{dec}}$. This two-stage operation yields the updated state $s_t$, context-aware image tokens $F_t'$, and pose-related output tokens $z_t'$, as formulated below:
\begin{equation}
\begin{aligned}
F_t = f_{\text{enc}}(I_t), \quad
[z_t', F_t'], s_t = f_{\text{dec}}([z, F_t], s_{t-1})
\end{aligned}
\label{eq:cut3r_processing_stages}
\end{equation}
Following this, dedicated prediction heads utilize these enriched tokens ($F_t'$ and $z_t'$) to output 3D point maps $P_{\text{map}t}$ and the relative camera pose (ego-motion) $\mathcal{T}_t$. Note that CUT3R produces metric-scale point maps and camera pose estimates, rather than outputs in a normalized scale~\cite{wang2024dust3r,wang2025vggt}, which reduces the difficulty of spatial instruction alignment.
Rather than directly fusing explicit point clouds, which can be sparse, non-uniformly sized across frames, and challenging to encode, we leverage the implicit latent representations. Specifically, the enriched spatial tokens $F_t'$ and the camera view tokens $z_t'$ (derived from $f_{\text{dec}}$) collectively serve as our rich \textit{3D reconstructive tokens}, compactly encoding the observed 3D geometry and camera perspective. To ensure consistency in the number of tokens generated by the visual and these reconstructive (spatial) encoders, input images are resized to a standard size (e.g., $432 \times 432$ pixels). During subsequent model training, the weights of both the pre-trained visual encoder and the CUT3R-based spatial encoders ($f_{\text{enc}}$, $f_{\text{dec}}$) are frozen.

\vspace{-3mm}
\paragraph{Spatial-visual View Fusion.}
VLM-3R integrates 3D geometric information using specialized 3D reconstructive tokens derived from our 3D tokenization process (Sec.~\ref{sec:3dtoken}). Specifically, these tokens consist of enriched spatial tokens $F_t'$ and camera view tokens $z_t'$, which are concatenated to form a unified 3D representation, $Z_{3D} = \text{Concat}(F_t', z_t')$. We then fuse this 3D representation with the VLM's native visual tokens $H_v$ (extracted from video frames by a pre-trained visual encoder, e.g., CLIP ViT) through a cross-attention mechanism, where $H_v$ serves as queries and $Z_{3D}$ provides keys and values. The cross-attention output is computed as
\begin{equation}
H_{\text{attn}} = \text{softmax}\left(\frac{(H_v W_Q)(Z_{3D} W_K)^T}{\sqrt{d_k}}\right)(Z_{3D} W_V),
\label{eq:cross_attention_fusion}
\end{equation}
where $W_Q$, $W_K$, and $W_V$ are learnable projection matrices, and $d_k$ is the key dimension. To preserve the original visual appearance information while injecting 3D priors, we apply a residual connection and obtain the enriched visual representation
\begin{equation}
H_v' = H_v + H_{\text{attn}}.
\end{equation}
The fused tokens $H_v'$ are then passed through a two-layer projector, following the design used in models such as LLaVA-Next-Video~\cite{zhang2024llavanextvideo}, to align them with the input space of the LMM backbone. The resulting 3D-aware visual tokens are concatenated with the language instruction tokens $H_{\text{instruct}}$ (e.g., $[H_v'; H_{\text{instruct}}]$) and fed into the transformer backbone. End-to-end supervised fine-tuning on our curated reconstructive-instructional data enables the model to jointly reason over visual appearance, 3D geometric priors, and camera perspective, thereby improving 3D spatial understanding.


\vspace{-3mm}
\paragraph{Training Objective and Fine-tuning Strategy}
For training VLM-3R, we adopt the same learning objective as LLaVA-NeXT-Video. To achieve efficient adaptation, we employ Low-Rank Adaptation (LoRA)~\cite{hu2022lora} for fine-tuning for VLM, and involves updating parameters within 3D fusion attention block and the projection layers.

%% file: sec/5_experiment.tex
\section{Experiments}
\subsection{Implementation Details}

\begin{table*}[t]
    \captionsetup{type=table}
    \centering
    \fontsize{4.2pt}{4.0pt}\selectfont
    \setlength\tabcolsep{2.8pt}
    \renewcommand{\arraystretch}{1.15}

    \resizebox{0.7\textwidth}{!}{
    \begin{tabular}{r|cc|cccccccc}
    & & &
    \rotatebox{75}{Obj. Count} &
    \rotatebox{75}{Abs. Dist.} &
    \rotatebox{75}{Obj. Size} &
    \rotatebox{75}{Room Size} &
    \rotatebox{75}{Rel. Dist.} &
    \rotatebox{75}{Rel. Dir.} &
    \rotatebox{75}{Route Plan} &
    \rotatebox{75}{Appr. Order} \\
    Methods & Rank & Avg. &
    \multicolumn{4}{c}{\cellcolor{orange!10}Numerical Answer} &
    \multicolumn{4}{c}{\cellcolor{yellow!10}Multiple-Choice Answer} \\
    \hline
    \rowcolor{navyblue!5}
    \multicolumn{11}{l}{\textcolor{black}{\textit{Baseline}}} \\
    Chance Level (Random) & - & - & - & - & - & - & 25.0 & 36.1 & 28.3 & 25.0 \\
    Chance Level (Frequency) & - & 34.0 & 62.1 & 32.0 & 29.9 & 33.1 & 25.1 & 47.9 & 28.4 & 25.2 \\
    \hline
    \rowcolor{navyblue!5}
    \multicolumn{11}{l}{\textcolor{black}{\textit{Proprietary Models (API)}}} \\
    Gemini-2.5 Pro & 1 & 51.5 & 43.8 & 34.9 & 64.3 & 42.8 & 61.1 & 47.8 & 45.9 & 71.3 \\
    GPT-4o         & 2 & 34.0 & 46.2 & 5.3  & 43.8 & 38.2 & 37.0 & 41.3 & 31.5 & 28.5 \\
    \hline
    \rowcolor{navyblue!5}
    \multicolumn{11}{l}{\textcolor{black}{\textit{Open-source VLMs}}} \\
    LongVILA-8B          & 13 & 21.6 & 29.1 & 9.1  & 16.7 & 0.0  & 29.6 & 30.7 & 32.5 & 25.5 \\
    VILA-1.5-8B          & 12 & 28.9 & 17.4 & 21.8 & 50.3 & 18.8 & 32.1 & 34.8 & 31.0 & 24.8 \\
    Qwen2.5-VL-7B        & 11 & 29.2 & 25.2 & 10.9 & 35.8 & 29.2 & 38.7 & 37.5 & 29.4 & 26.7 \\
    LongVA-7B            & 11 & 29.2 & 38.0 & 16.6 & 38.9 & 22.2 & 33.1 & 43.3 & 25.4 & 15.7 \\
    VILA-1.5-40B         & 10 & 31.2 & 22.4 & 24.8 & 48.7 & 22.7 & 40.5 & 25.7 & 31.5 & 32.9 \\
    LLaVA-OneVision-7B   & 9  & 32.4 & 47.7 & 20.2 & 47.4 & 12.3 & 42.5 & 35.2 & 29.4 & 24.4 \\
    LLaVA-NeXT-Video-7B  & 8  & 35.6 & 48.5 & 14.0 & 47.8 & 24.2 & 43.5 & 42.4 & 34.0 & 30.6 \\
    LLaVA-OneVision-72B  & 7  & 40.2 & 43.5 & 23.9 & 57.6 & 37.5 & 42.5 & 39.9 & 32.5 & \cellcolor{oai-gray-300}{44.6} \\
    LLaVA-NeXT-Video-72B & 6  & 40.9 & 48.9 & 22.8 & 57.4 & 35.3 & 42.4 & 36.7 & 35.0 & \cellcolor{oai-gray-600}{48.6} \\
    Spatial-MLLM-4B      & 5  & 47.0 & 65.3 & 34.8 & 63.1 & 45.1 & 41.3 & 46.2 & 33.5 & 46.3 \\
    VG-LLM-4B            & 4  & 47.3 & 66.0 & 37.8 & 55.2 & 59.2 & 44.6 & 45.6 & 33.5 & 36.4 \\
    LLaVA-One-Vision-7B \textit{(Finetuned)}  & 3  & 55.8 & 68.3 & \cellcolor{oai-gray-300}{45.3} & 66.5 & 59.5 & 59.7 & 65.9 & \cellcolor{oai-gray-300}{41.7} & 39.4 \\
    LLaVA-Next-Video-7B \textit{(Finetuned)}  & 2  & 57.7 & \cellcolor{oai-gray-600}{70.6} & 43.6 & \cellcolor{oai-gray-600}{70.8} & \cellcolor{oai-gray-300}{63.7} & \cellcolor{oai-gray-300}{64.9} & \cellcolor{oai-gray-300}{68.9} & 40.7 & 38.5 \\
    \textbf{VLM-3R (7B)} & \cellcolor{oai-green-600}{1} & 60.9 &
       \cellcolor{oai-gray-300}{70.2} & \cellcolor{oai-gray-600}{49.4} &
       \cellcolor{oai-gray-300}{69.2} & \cellcolor{oai-gray-600}{67.1} &
       \cellcolor{oai-gray-600}{65.4} & \cellcolor{oai-gray-600}{80.5} &
       \cellcolor{oai-gray-600}{45.4} & 40.1 \\
    \hline
    \end{tabular}
    }
    \caption{\textbf{Evaluations on VSI-Bench.}
    Open-source models, including finetuned variants, are jointly ranked by Avg. Models marked with \textit{(Finetuned)} are further trained using our 200K spatial reasoning QA pairs.
    \textbf{VLM-3R} ranks first among open-source VLMs, demonstrating the effectiveness of its reconstructive instruction tuning.
    Within the combined open-source block, \colorbox{oai-gray-600}{dark gray} marks the best score and \colorbox{oai-gray-300}{light gray} marks the second best. Tied Avg values share the same rank.}
    \label{tab:vsibench}
\end{table*}

\paragraph{Baselines and Setup.}
We follow prior Visual–Spatial intelligence evaluations~\cite{yang2024thinking} and compare against a broad set of video LMMs, including proprietary systems (e.g., Gemini~\cite{team2024gemini}, GPT-4o~\cite{hurst2024gpto}) and strong open-source models (e.g., ViLA~\cite{lin2024vila}, LLaVA-OneVision~\cite{li2024llavaov}, LLaVA-NeXT-Video~\cite{zhang2024llavanextvideo}, Qwen2.5-VL~\cite{bai2025qwen2}, Spatial-MLLM~\cite{wu2025spatial}, VG-LLM~\cite{zheng2025learning}). 
Our model is fine-tuned using LoRA~\cite{hu2022lora} (rank 128, scale 256), keeping visual and spatial encoders frozen except for one alignment block. Training is performed for one epoch on NVIDIA H200 GPUs.

\subsection{Benchmarks.}
We evaluate VLM-3R on a diverse set of benchmarks that cover spatial, temporal, and general understanding. 
For 3D spatial perception, we use VSI-Bench~\cite{yang2024thinking}; 
for temporal reasoning, VS\textit{Temporal}I-Bench measures camera motion and camera–object dynamics; 
and for online spatio-temporal comprehension, we adopt OST-Bench~\cite{lin2025ost}. 
General understanding is evaluated on the video benchmark Video-MME~\cite{fu2025video}.
and the image benchmark VQAv2~\cite{goyal2017making}.


\vspace{-4mm}
\paragraph{Evaluation on VSI-Bench}
We evaluate on VSI-Bench~\cite{yang2024thinking}, which contains over 5{,}000 QA pairs from egocentric videos in ScanNet, ScanNet++, and ARKitScenes. 
The benchmark includes Multiple-Choice (MCA) and Numerical Answer (NA) formats; we follow the official protocol using mean accuracy for MCA and relative accuracy for NA. 
VSI-Bench spans eight spatial reasoning tasks, and results for each are reported in Table~\ref{tab:vsibench}.

We observe substantial gains across multiple task types when comparing the 2D-only baseline to our full VLM-3R model.
For example, performance on \textit{Absolute Distance} increases from \textbf{20.2} to \textbf{49.4}, and \textit{Room Size} improves from \textbf{12.3} to \textbf{67.1}.
The improvement on \textit{Relative Direction} is particularly notable, rising from \textbf{42.4} to \textbf{80.5} and representing one of the largest gains across the benchmark.
These results demonstrate that our carefully designed architecture and spatially oriented training data work together to substantially enhance spatial understanding in VLM-3R.

\vspace{-4mm}
\paragraph{Evaluation on VSTI-Bench}
To evaluate VLMs' capabilities in spatial reasoning over time, we report performance on VSTI-Bench, which measures the understanding of \textbf{\textit{temporally}} evolving dynamics. Table~\ref{tab:vstibench} details tasks that require interpreting complex events in monocular videos, such as tracking static objects under moving cameras and reasoning about their interactions across space and time. On VSTI-Bench, VLM-3R shows strong understanding of both spatial context and temporal motion, enabling it to answer questions and make inferences about video content. These results validate that incorporating view tokens from the geometry encoder helps disentangle camera movement from object-centric spatial relationships.

\begin{table}[!ht]
    \centering
    \vspace{-3mm}
    \scriptsize
    \setlength\tabcolsep{2pt}
    \resizebox{\columnwidth}{!}{
    \begin{tabular}{r|cc|ccccc}
    & & & 
    \rotatebox{75}{Cam-Obj Abs. Dist.} & 
    \rotatebox{75}{Cam. Displace.} &     
    \rotatebox{75}{Cam. Mov. Dir.} &     
    \rotatebox{75}{Obj-Obj Rel. Pos.} &  
    \rotatebox{75}{Cam-Obj Rel. Dist.} \\ 

    Methods & Rank & Avg. &
    \multicolumn{2}{c}{\cellcolor{orange!10}Numerical Answer} & 
    \multicolumn{3}{c}{\cellcolor{yellow!10}Multiple-Choice Answer} \\ 
    \hline

    \rowcolor{navyblue!5}
    \multicolumn{8}{l}{\textcolor{black}{\textit{Baseline}}} \\
    Chance Level (Random) & - & - & - & - & 36.1 & 50.0 & 36.1 \\
    Chance Level (Frequency) & - & 27.4 & 5.4 & 6.2 & 40.7 & 52.2 & 32.4 \\
    \hline

    \rowcolor{navyblue!5}
    \multicolumn{8}{l}{\textcolor{black}{\textit{Human Performance}}} \\
    \dag Human Level & - & 77.0 & 51.4 & 46.8 & 95.1 & 97.5 & 94.3 \\
    \hline

    \rowcolor{navyblue!5}
    \multicolumn{8}{l}{\textcolor{black}{\textit{Proprietary Models (API)}}} \\
    \textbf{Gemini-2.5 Pro} & \cellcolor{oai-green-600}{1} & \textbf{42.4} & 29.2 & 8.0 & 30.0 & 84.0 & 60.7 \\
    GPT-4o & \cellcolor{oai-green-400}{2} & 38.2 & 29.5 & 23.4 & 37.3 & 58.1 & 42.5 \\
    \hline

    \rowcolor{navyblue!5}
    \multicolumn{8}{l}{\textcolor{black}{\textit{Open-sourced VLMs}}} \\

    LLaVA-NeXT-Video-7B & 5 & 40.0 & 28.2 & 1.8 & \cellcolor{oai-gray-300}{49.8} & 64.7 & \cellcolor{oai-gray-300}{55.6} \\
    LLaVA-OneVision-7B & 4 & 41.7 & 29.9 & 19.3 & 47.5 & 62.1 & 49.8 \\
    LongVA-7B & 10 & 32.3 & 13.5 & 5.1 & 43.7 & 57.9 & 41.2 \\
    InternVL2-8B & \cellcolor{oai-green-200}{3} & 43.5 & \cellcolor{oai-gray-300}{32.9} & 13.5 & 48.0 & 68.0 & 55.0 \\
    LongVILA-8B & 11 & 30.5 & 20.0 & 11.6 & 35.4 & 52.3 & 33.4 \\
    VILA-1.5-8B & 8 & 37.3 & 30.1 & 27.3 & 42.2 & 50.4 & 36.7 \\
    \hline

    VILA-1.5-40B & 6 & 38.2 & 28.2 & 15.7 & 28.8 & 65.4 & 53.0 \\

    LLaVA-NeXT-Video-72B & \cellcolor{oai-green-400}{2} & \cellcolor{oai-gray-300}{44.0} & 32.3 & 10.5 & 48.1 & \cellcolor{oai-gray-300}{78.3} & 50.9 \\
    \hline

    \textbf{VLM-3R (7B)} & \cellcolor{oai-green-600}{1} & \cellcolor{oai-gray-600}{58.8} 
    & \cellcolor{oai-gray-600}{39.4} & \cellcolor{oai-gray-600}{39.6} 
    & \cellcolor{oai-gray-600}{60.6} & \cellcolor{oai-gray-600}{86.5} & \cellcolor{oai-gray-600}{68.6} \\
    \hline
    \end{tabular}
    }

    \caption{\textbf{Evaluations on VS\textit{Temporal}I-Bench.}
    VLM-3R achieves leading performance, demonstrating strong spatio-temporal reasoning and robust understanding of camera dynamics.}
    \label{tab:vstibench}
\end{table}

\vspace{-4mm}
\paragraph{Evaluation on ScanQA and SQA3D}
ScanQA~\cite{azuma2022scanqa} and SQA3D~\cite{ma2022sqa3d} are 3D QA benchmarks. 
For ScanQA, we follow prior work and evaluate on its 4{,}675-question validation split using standard language metrics (CIDEr, BLEU, METEOR, ROUGE-L). 
SQA3D contains 3{,}519 test QA pairs that require spatial reasoning in 3D scenes, and we report exact-match accuracy (EM) and its refined variant EM-R.
Following the Spatial-MLLM~\cite{wu2025spatial} protocol, we train VLM-3R on the combined dataset, including the official training splits of ScanQA and SQA3D, to ensure a standardized evaluation setup.
As shown in Table~\ref{tab:scanqa_sqa3d}, VLM-3R outperforms video-only LLMs and is competitive with 2.5D/3D models like Video-3D LLM~\cite{zheng2025video} and LLaVA-3D~\cite{zhu2024llava}, which require ground-truth depth. Despite using only monocular video, VLM-3R achieves competitive performance with depth-supervised baselines, validating the effectiveness of our spatial–visual–language reasoning and demonstrating strong generalization to real 3D QA benchmarks.


\begin{table}[t]
\centering
\scriptsize
\setlength\tabcolsep{1.8pt}
\renewcommand{\arraystretch}{1.08}
\resizebox{0.47\textwidth}{!}{%
\begin{tabular}{l|ccccc|cc|m{9mm}}
\toprule
\textbf{Methods} &
\multicolumn{5}{c|}{\textbf{ScanQA (val)}} &
\multicolumn{2}{c|}{\textbf{SQA3D (test)}} &
\textbf{\shortstack[c]{Video\\Input\\Only}} \\
\cmidrule(lr){2-6} \cmidrule(lr){7-8}
& \textbf{B1} & \textbf{B4} & \textbf{M} & \textbf{RL} & \textbf{C} & \textbf{EM1} & \textbf{EMR1} &  \\
\midrule
\rowcolor{gray!10}
\multicolumn{9}{l}{\textit{Task-Specific Models}} \\
ScanQA~\cite{azuma2022scanqa} & 30.2 & 10.1 & 13.1 & 33.3 & 64.9 & 47.2 & -- & \xmark \\
SQA3D~\cite{ma2022sqa3d}   & 30.5 & 11.2 & 13.5 & 34.5 & --   & 46.6 & -- & \xmark \\
3D-Vista~\cite{zhu20233d} & -- & -- & 13.9 & 35.7 & -- & 48.5 & -- & \xmark \\
\midrule
\rowcolor{gray!10}
\multicolumn{9}{l}{\textit{3D/2.5D-Input Models}} \\
3D-LLM~\cite{hong20233d}         & 39.3 & 12.0 & 14.5 & 35.7 & 69.4 & --   & --   & \xmark \\
LL3DA~\cite{chen2024ll3da}           & --   & 13.5 & 15.9 & 37.3 & 76.8 & --   & --   & \xmark \\
Chat-Scene~\cite{huang2024chat}  & 43.2 & 14.3 & 18.0 & 41.6 & 87.7 & 54.6 & 57.5 & \xmark \\
3D-LLaVA~\cite{deng20253d}      & --   & 17.1 & 18.4 & 43.1 & 92.6 & 54.5 & 56.6 & \xmark \\
Video-3D LLM~\cite{zheng2024video}& 47.1 & 16.2 & 19.8 & 49.0 & 102.1& 58.6 & --   & \xmark \\
\midrule
\rowcolor{gray!10}
\multicolumn{9}{l}{\textit{Video-Input Models}} \\
Qwen2.5-VL-3B~\cite{bai2025qwen2}   & 22.5 &  3.8 &  9.7 & 25.4 & 47.4 & 43.4 & 45.9 & \cmark \\
Qwen2.5-VL-7B~\cite{bai2025qwen2}   & 27.8 &  8.0 & 11.4 & 29.3 & 53.9 & 46.5 & 49.8 & \cmark \\
Qwen2.5-VL-72B~\cite{bai2025qwen2}  & 26.8 & 12.0 & 13.0 & 35.2 & 66.9 & 47.0 & 50.9 & \cmark \\
LLaVA-Video-7B~\cite{zhang2024video}& 39.7 &  3.1 & 17.7 & 44.6 & 88.7 & 48.5 & --   & \cmark \\
Oryx-34B~\cite{liu2024oryx}            & 38.0 & --   & 15.0 & 37.3 & 72.3 & 50.9 & --   & \cmark \\
Spatial-MLLM-4B~\cite{wu2025spatial} & 44.4 & 14.8 & 18.4 & 45.0 & 91.8 & 55.9 & 58.7 & \cmark \\
\textbf{VLM-3R (Ours)}          & \textbf{46.2} & \textbf{15.5} & \textbf{19.7} & \textbf{49.1} & \textbf{101.9} & \textbf{60.7} & \textbf{63.4} & \cmark \\
\bottomrule
\end{tabular}}
\caption{\textbf{Evaluation on ScanQA and SQA3D.}
Comparison across task-specific, 3D/2.5D-input, and video-input models.}
\label{tab:scanqa_sqa3d}
\end{table}

\begin{table}[t]
    \centering
    \small
    \setlength\tabcolsep{6pt}
    \renewcommand{\arraystretch}{1.15}
    \resizebox{0.48\textwidth}{!}{
    \begin{tabular}{lccc}
        \toprule
        \textbf{Method} &
        \makecell[c]{\textbf{Video-MME:}\\\textbf{Overall}} &
        \makecell[c]{\textbf{VQA v2:}\\\textbf{Exact Match}} &
        \makecell[c]{\textbf{Video-MME:}\\\textbf{Spatial Perception}} \\
        \midrule
        \rowcolor{oai-gray-200}
        \textbf{VLM-3R (Ours)} & 59.9 & 52.57 & \textbf{70.4} \\
        LLaVA-NeXT-Video & \textbf{62.7} & \textbf{54.63} & 66.7 \\
        \bottomrule
    \end{tabular}
    }
    \vspace{-3mm}
    \caption{\textbf{General-purpose and spatial perception understanding.}
    VLM-3R retains strong video- and image-level understanding while achieving higher spatial perception accuracy.}
    \label{tab:general_understanding}
    \vspace{-3mm}
\end{table}

\paragraph{Effect of Mixing General Video Data.}
To enhance general video understanding, we follow the \textit{LLaVA-3D}~\cite{zhu2024llava} data-mixing strategy by adding $30$k general videos from \textit{LLaVA-Video}~\cite{zhang2024video} to the $200$k domain-specific \textit{VSI-Bench}. 
As shown in Table~\ref{tab:data_mixing}, mixing general video data lifts \textit{Video-MME} performance from \textbf{59.9\%} to \textbf{62.1\%} ($+2.2$ pp), while the mixed model remains within \textbf{1pp} of the original \textit{LLaVA-NeXT-Video-7B}. 
This phenomenon is consistent with the observation in \textit{LLaVA-3D}, where mixing additional domain-specific data keeps the general video QA performance within 1pp of the original model.
We note that the officially reported \textit{LLaVA-NeXT-Video-7B} score is \textbf{63.3\%}, whereas our reproduction yields \textbf{62.7\%}, which we attribute to environment differences.


\begin{table}[t]
\centering
\footnotesize
\setlength{\tabcolsep}{4pt}
\renewcommand{\arraystretch}{1.05}
\begin{tabular}{lc}
\toprule
\textbf{Training Strategy} & \textbf{Video-MME Overall (\%)} \\
\midrule
LLaVA-NeXT-Video-7B & 62.7 \\
VSI-only & 59.9 \\
VSI + LLaVA & 62.1 \\
\bottomrule
\end{tabular}
\vspace{-3mm}
\caption{\textbf{Effect of data-mixing strategy on general video understanding.}
\textbf{VSI-only}: trained solely on the domain-specific \textit{VSI-Bench} (200k). 
\textbf{VSI + LLaVA}: mixed-domain training that adds 30k general video samples from \textit{LLaVA-Video}. 
The data-mixing strategy restores \textit{Video-MME} performance from $59.9\%$ to $62.1\%$, 
showing improved generalization without architectural change.}
\label{tab:data_mixing}
\vspace{-3mm}
\end{table}

\paragraph{Generalization to Spatio-Temporal Understanding}
\label{sec:ostbench}
OST-Bench~\cite{lin2025ost} provides a realistic evaluation setup, requiring models to perform online, temporally grounded reasoning from actively exploring agents, rather than relying on static or pre-recorded inputs. Although VLM-3R is trained solely on static indoor scenes, it generalizes well to this online embodied exploration setting. As shown in Table~\ref{tab:ostbench}, we compare three spatially tuned models (Spatial-MLLM, LLaVA-3D, and VLM-3R) against their respective base models. VLM-3R outperforms its base model across many categories, particularly in \textit{Agent State}, \textit{Agent–Object Spatial Relationship}, and \textit{Estimation} tasks, demonstrating stronger robustness and cross-domain generalization in this challenging online evaluation.

\paragraph{General-Purpose Visual Understanding.}
We evaluate VLM-3R's generalization on video and image understanding benchmarks, including Video-MME~\cite{fu2025video} and VQA v2~\cite{goyal2017making}. 
Video-MME provides full-spectrum evaluation for MLLMs, spanning six domains and 30 categories with multimodal inputs. 
VQA v2 is a widely used balanced visual benchmark designed to reduce language priors. 
As shown in Table~\ref{tab:general_understanding}, despite not being trained on generic video QA data, VLM-3R performs comparably to \textit{LLaVA-NeXT-Video}. 
Notably, VLM-3R improves \textit{Spatial Perception} by +\textbf{3.7}\% on Video-MME. 
These results indicate that our spatial–visual fusion strategy generalizes well to both 2D and video comprehension while enhancing spatial perception.

\begin{table}[t]
\centering
\footnotesize
\setlength{\tabcolsep}{4pt}
\renewcommand{\arraystretch}{1.1}
\begin{tabular}{lcc}
\toprule
\textbf{Method} & \textbf{Rank} & \textbf{Avg. (\%)} \\
\midrule
\rowcolor{navyblue!10}
\textbf{VLM-3R (Full / 2D–3D fusion)} & \textbf{1} & \textbf{60.90} \\
VLM-3R w/o Spatial Tok. & 2 & 59.46 \\
VLM-3R w/o View Tok. & 3 & 59.09 \\
VLM-3R (2D–2D fusion) & 4 & 58.12 \\
VLM-3R (Explicit Points Fusion) & 5 & 57.87 \\
LLaVA-NeXT-Video ft (w/o C\&G Tok.) & 6 & 57.74 \\
\bottomrule
\end{tabular}
\caption{\textbf{Ablation on VLM-3R components and fusion strategies.}
We report only overall performance (\textit{Avg.}) on \textit{VSI-Bench}.
Removing spatial or view tokens slightly reduces performance,
while full 2D–3D fusion yields the best overall result.}
\label{tab:vlm3r_ablation}
\end{table}

\begin{table}[t]
\centering
\setlength\tabcolsep{2.8pt} 
\renewcommand{\arraystretch}{1.15}
\vspace{1mm}
\resizebox{\linewidth}{!}{ 
\begin{tabular}{l|cccccccc}
\toprule
\textbf{Method} & \textbf{Overall} & \textbf{JUD.} & \textbf{EST.} & \textbf{CNT.} & \textbf{TEMP.} & \textbf{A. State} & \textbf{A. Info} & \textbf{AO.} \\
\midrule
(base) QwenVL2.5-3B & 34.8 & 47.9 & 18.7 & 59.4 & 19.8 & 34.2 & 47.5 & 25.7 \\
Spatial-MLLM        & 26.8 & 37.3 & 21.9 & 29.5 & 15.3 & 25.5 & 39.4 & 20.9 \\
\midrule
(base) LLaVA-Video-7B & 39.3 & 52.8 & 16.1 & 63.1 & 33.8 & 33.5 & \textbf{58.3} & 28.8 \\
\textbf{VLM-3R}       & \textbf{42.9} & \textbf{55.1} & \textbf{28.3} & 49.6 & \textbf{36.0} & \textbf{39.9} & 58.1 & \textbf{34.4} \\
LLaVA-3D              & 30.1 & 46.1 & 5.9  & 13.5 & 36.3 & 29.7 & 38.4 & 26.3 \\
\bottomrule
\end{tabular}
}
\caption{\textbf{Comparison between spatially grounded models and their base models on OST-Bench.} 
\textbf{VLM-3R} consistently outperforms its base (LLaVA-Video-7B) across all categories, 
particularly in \textit{Agent State}, \textit{Agent–Object Spatial Relationship}, and \textit{Estimation} tasks.}
\label{tab:ostbench}
\end{table}

\subsection{Ablation Studies}
We evaluate each component's effectiveness. All variants are trained on our 200K spatial data (Sec.~\ref{sec:data-vsi}) and evaluated on VSI-Bench, following the VLM-3R setup (Table~\ref{tab:vlm3r_ablation}).

\vspace{-3mm}
\paragraph{Spatial Tokens Fusion}
Removing spatial tokens (\texttt{w/o Spatial Token}) leads to a clear performance drop in structure-dependent tasks, with the overall score decreasing from 60.90 to 59.46, confirming their role in modeling 3D layout and depth relations.


\vspace{-3mm}
\paragraph{View Tokens Fusion}
Without view tokens (\texttt{w/o View Token}), direction-sensitive tasks degrade, with the overall score dropping to 50.09, showing that pose encoding is important for maintaining an egocentric spatial frame.

\vspace{-3mm}
\paragraph{2D–Only Fusion}
The 2D–2D fusion variant (\texttt{2D–2D Fusion}) omits geometric cues and yields lower accuracy (average 58.12 vs. 60.90), especially in reasoning tasks, highlighting the necessity of 3D spatial integration.

\vspace{-3mm}
\paragraph{Token vs. Explicit Points Fusion}
Directly fusing point clouds (\texttt{Explicit Points Fusion}) performs worse overall (57.87 vs. 60.90), indicating that token-level fusion offers more stable cross-modal reasoning and avoids sparsity issues in explicit 3D data.

\vspace{-3mm}
\paragraph{Overall 3D Fusion Framework}
Compared to \texttt{LLaVA-NeXT-Video ft (w/o C\&G Tok.)}, our full model achieves a 3.2-point gain (60.90 vs. 57.74), validating the proposed 3D fusion. 
Slightly lower \textit{Object Size} accuracy (69.15 vs. 70.82) suggests room for improvement in monocular 3D reconstruction quality.

%% file: sec/6_conclusion.tex
\section{Conclusion}

In this work, we introduce VLM-3R, a novel framework that significantly enhances Vision-Language Models (VLMs) through reconstructive instruction tuning. By leveraging over 200K curated training instances and a Spatial-Visual-View fusion module that integrates geometric information with camera view context, VLM-3R enables robust 3D spatial understanding directly from monocular video, eliminating the need for depth sensors or pre-computed 3D maps and helping disentangle object semantics from egocentric camera motion. To address a critical gap in current evaluations, we also introduce a new benchmark with approximately 138.6K question–answer pairs, specifically designed to assess the often overlooked temporal reasoning abilities of these models, while highlighting that their broader applicability and ultimate performance remain closely tied to the challenges of large-scale 4D data collection and the accuracy of end-to-end 3D reconstruction. As a foundational step, our current datasets prioritize static indoor scenes, leaving the exploration of dynamic and extreme environments for future work.


\noindent \textbf{Acknowledgments.}
This research has been supported by the Vista GPU Cluster through the Center for Generative AI (CGAI) and the Texas Advanced Computing Center (TACC) at the University of Texas at Austin.
ZF was supported by Gifts from Meta.

%% file: sec/X_suppl.tex
\clearpage
\setcounter{section}{0}
\setcounter{figure}{0}
\setcounter{table}{0}
\maketitlesupplementary

\renewcommand\thesection{\Alph{section}} 
\renewcommand\thesubsection{\thesection.\arabic{subsection}} 
\renewcommand\thefigure{\Alph{figure}} 
\renewcommand\thetable{\Alph{table}} 

\section{Additional Implementation Details}
\label{sec:appendix_impl_details}

\subsection{VLM-3R Architecture Specifics}
\paragraph{Spatial Encoder Configuration} 
For 3D scene reconstruction/understanding from monocular video, our VLM-3R utilizes a feed-forward (end-to-end) dense 3D reconstruction model as its spatial encoder. This type of encoder directly maps sequences of input RGB images $\{I_i\}_{i=1}^N$, where each $I_i \in \mathbb{R}^{3 \times H \times W}$, to 3D representations, including estimated camera intrinsics, extrinsics, and point maps that associate pixel information with 3D coordinates. To circumvent the extensive computational costs of using global point cloud and scaling difficulties across different scenes, we propose aligning the implicit encoder tokens instead.

We employ the CUT3R model~\cite{wang2025continuous} to process sequences of input images. Given multiple images, such as frames $\{I_t\}$ from a monocular video stream, CUT3R directly outputs corresponding dense 3D point maps (e.g., $\hat{X}^{\text{world}}_t$) and relative camera poses ($\hat{P}_t$) for each view.

\paragraph{Spatial-Visual-View Fusion Design}
VLM-3R employs a Spatial-Visual-View Fusion stage to integrate diverse multimodal inputs, as depicted in our overall architecture (e.g., Figure~3). This stage utilizes 3D reconstructive tokens from our spatial encoder~\cite{wang2025continuous}—specifically, geometry tokens $F_t'$ (encoding scene structure, dimension $729 \times 768$) and camera view tokens $z_t'$ (capturing global camera motion, dimension $1 \times 768$). These 3D tokens are fused with 2D appearance-based visual tokens $H_v$ (dimension $729 \times 1152$) derived from a pre-trained visual encoder, such as a CLIP ViT. Initially, the $F_t'$ and $z_t'$ tokens are concatenated to form a unified 3D representation, $Z_{3D} = \text{Concat}(F_t', z_t')$. This $Z_{3D}$ representation is then processed by a one-layer cross-attention block where it interacts with the visual tokens $H_v$ (e.g., $H_v$ as queries attending to $Z_{3D}$ as keys and values). The output of this attention stage, let's call it $H_{attn}$, is then residually connected with the original visual tokens $H_v$ to form an enriched representation $H_v' = H_v + H_{attn}$. This enriched $H_v'$ subsequently passes through a two-layer MLP projector, which transforms the feature dimensions (e.g., from $1152$ to $3584$, and then to $3584$) for effective alignment with the Large Multimodal Model (LMM). Finally, these fused and projected 3D-aware visual features, which constitute the final visual input to the LMM, are concatenated with language instruction tokens for processing by the LMM backbone.

\subsection{Training Details} 
\label{sec:appendix_training_details}
The VLM-3R model was trained using the following configuration and hyperparameters.
\paragraph{Base Model and Pretraining}
The training initialized from the \texttt{LLaVA-Video-7B-Qwen2} checkpoint. The vision tower used was \texttt{google/siglip-so400m-patch14-384}.

\paragraph{Hardware and Distributed Training Setup}
Training was conducted using a distributed setup. The specific training run utilized 16 H200 GPUs and lasted for approximately 5 hours. For distributed training, \texttt{NUM\_GPUS\_PER\_NODE} was set to 1, and the master address and port were dynamically determined using SLURM environment variables. The \texttt{accelerate} library with \texttt{torchrun} was used for launching distributed training, employing the DeepSpeed ZeRO stage 2 optimization (\texttt{scripts/zero2.json}). CPU affinity was set to 1 for the accelerate launcher.

\paragraph{Key Hyperparameters}
\begin{itemize}
    \item \textbf{LoRA Configuration}:
    \begin{itemize}
        \item \textbf{Enable LoRA}: \texttt{True} (\texttt{--lora\_enable True}).
        \item \textbf{LoRA rank}: \texttt{128} (\texttt{--lora\_r 128}).
        \item \textbf{LoRA alpha}: \texttt{256} (\texttt{--lora\_alpha 256}).
    \end{itemize}
    \item \textbf{Training Epochs}: The training was set for 5 epochs (\texttt{--num\_train\_epochs \$NUM\_TRAIN\_EPOCHS}), but concluded after the first epoch.
    \item \textbf{Batch Size}:
    \begin{itemize}
        \item \textbf{Per device training batch size}: \texttt{1} (\texttt{--per\_device\_train\_batch\_size 1}).
    \end{itemize}
    \item \textbf{Gradient accumulation steps}: \texttt{8} (\texttt{--gradient\_accumulation\_steps \$GRADIENT\_ACCUMULATION\_STEPS}).
    \item \textbf{Learning rate}: $2 \times 10^{-5}$ (\texttt{--learning\_rate 2e-5}).
    \item \textbf{Optimizer and Scheduler}:
    \begin{itemize}
        \item \textbf{Weight decay}: \texttt{0.0} (\texttt{--weight\_decay 0.}).
        \item \textbf{Warmup ratio}: \texttt{0.03} (\texttt{--warmup\_ratio 0.03}).
        \item \textbf{LR scheduler type}: \texttt{cosine} (\texttt{--lr\_scheduler\_type "cosine"}).
    \end{itemize}
    \item \textbf{Precision}: BF16 was enabled (\texttt{--bf16 True}). TF32 was also enabled (\texttt{--tf32 True}).
    \item \textbf{Model maximum length}: \texttt{32768} tokens (\texttt{--model\_max\_length 32768}).
    \item \textbf{Gradient Checkpointing}: Enabled (\texttt{--gradient\_checkpointing True}).
\end{itemize}

\paragraph{Spatial Module Configuration}
\begin{itemize}
    \item \textbf{Spatial tower}: \texttt{CUT3R} (\texttt{--spatial\_tower "cut3r"}).
    \item \textbf{Spatial tower feature selection}: \texttt{all} (\texttt{--spatial\_tower\_select\_feature "all"}).
    \item \textbf{Spatial feature dimension}: \texttt{768} (\texttt{--spatial\_feature\_dim "768"}).
    \item \textbf{Fusion block}: \texttt{cross\_attention} (\texttt{--fusion\_block "cross\_attention"}).
    \item \textbf{Tunable Components}:
    \begin{itemize}
        \item \textbf{Tune spatial tower}: \texttt{False} (\texttt{--tune\_spatial\_tower False}).
        \item \textbf{Tune fusion block}: \texttt{True} (\texttt{--tune\_fusion\_block True}).
        \item \textbf{Tune MM MLP adapter}: \texttt{True} (\texttt{--tune\_mm\_mlp\_adapter True}).
    \end{itemize}
\end{itemize}

\section{Dataset Curation and Benchmark Design} 
\label{sec:appendix_datasets}
3D Reconstructive Instructional Tuning relies on large-scale Question-Answer (QA) pairs to fine-tune Large Multimodal Models (LMMs), enabling them to perform tasks related to spatial and temporal reasoning. We explain the construction of our 3D Reconstructive datasets, which are utilized for training models subsequently evaluated on VSI-Bench~\cite{yang2024thinking}. Furthermore, we detail the creation of the Visual-Spatial-Temporal Intelligence Benchmark (VSTI-Bench), a new resource containing comprehensive training and evaluation pairs specifically for spatial-temporal tasks.

\subsection{Scalable 3D Reconstructive Instructional QA Creation} 
    \label{sec:appendix_training_data_custom} 
\paragraph{Data Sources and Preprocessing Methodology} 
Our training dataset for 3D Reconstructive Instructional QA is generated using a methodology inspired by approaches such as those used in the VSI-Bench (Visual-Spatial Intelligence Benchmark). Data is sourced from established 3D scene datasets such as ScanNet~\cite{dai2017scannet}, ScanNet++~\cite{yeshwanth2023scannet++}, and ARKitScenes~\cite{baruch2021arkitscenes}, which are unified and processed. The core of the data preparation involves a detailed preprocessing pipeline to extract structured scene-level and frame-level metadata, similar to established practices.

The input data requirements for each scene include:
\begin{enumerate}
    \item \textbf{Point Cloud (PCD):} A 3D point cloud (e.g., from \texttt{.ply} files) with per-point semantic labels, instance labels, coordinates, and color.
    \item \textbf{Video:} An RGB video traversal of the scene.
    \item \textbf{Sampled Frame Data:} A collection of frames sampled from the video, including color images, depth maps, instance segmentation masks, and their corresponding 6DoF camera poses (4x4 transformation matrices) in the world coordinate system.
    \item \textbf{Camera Intrinsics:} Parameters like focal length ($f_x, f_y$) and principal point ($c_x, c_y$).
\end{enumerate}
This raw data undergoes a preprocessing pipeline that generates two primary types of structured metadata files in JSON format:
\begin{itemize}
    \item \textbf{{scene\_metadata.json}:} Contains scene-wide information, including overall scene dimensions, room center, counts of different object categories, and detailed 3D bounding boxes (center, size, orientation, instance ID) for every object instance within the scene.
    \item \textbf{{frame\_metadata.json}:} Contains frame-specific information for each scene, such as camera intrinsics, image dimensions, and for each sampled frame: its ID, paths to color/depth images, the camera pose, and 2D bounding boxes for visible object instances.
\end{itemize}
These metadata files are crucial for the subsequent automated generation of diverse QA pairs for our training dataset.

\paragraph{Spatio-Temporal Scene Graph Construction from Metadata} 
The generated \texttt{scene\_metadata.json} and \texttt{frame\_metadata.json} files effectively serve as a structured spatio-temporal scene graph. This representation includes precise 3D locations, sizes, and orientations of objects, their semantic and instance-level identities, and the dynamic viewpoint of the camera across multiple frames. Such detailed and organized metadata enables the querying of complex spatial relationships (e.g., object-object relations, object-camera distances) and spatiotemporal changes (e.g., camera movement, object appearance order). This structured understanding forms the foundation for the diverse QA tasks in our generated dataset.

\paragraph{Spatial QA Generation Methodology} 
The QA pairs for our training dataset are systematically generated using dedicated scripts for each task type, leveraging the aforementioned structured metadata. The generation logic for these tasks is the same as that used in VSI-Bench, and the data is intended for training models on various facets of visual-spatial intelligence.

\begin{itemize}
    \item \textbf{Configurational Tasks:} These assess understanding of spatial layout and inter-object relationships.
    \begin{itemize}
        \item \textit{Object Count:} Counting instances of a specific object category in a room (numerical answer; objects with single instances are excluded).
        \item \textit{Relative Distance:} Determining which of four candidate objects is closest in 3D space to a target object (multiple-choice answer).
        \item \textit{Relative Direction:} Identifying the direction of a query object relative to an observer at a specific position and orientation (multiple-choice answer, e.g., left/right/back).
        \item \textit{Route Plan:} Completing a sequence of navigation actions (Go forward, Turn) to reach a target object from a starting object (multiple-choice answer).
    \end{itemize}
    \item \textbf{Measurement Estimation Tasks:} These require quantitative estimation of spatial properties.
    \begin{itemize}
        \item \textit{Object Size:} Estimating the length of the longest dimension of a unique object instance in centimeters (numerical answer).
        \item \textit{Absolute Distance:} Estimating the direct Euclidean distance between the closest points of two specified objects in meters (numerical answer).
        \item \textit{Room Size:} Estimating the area of the room in square meters (numerical answer).
    \end{itemize}
    \item \textbf{Spatiotemporal Task:} This tests the processing of spatial information over time.
    \begin{itemize}
        \item \textit{Appearance Order:} Determining the first-appearance order of four object categories in the video sequence (multiple-choice answer).
    \end{itemize}
\end{itemize}

\begin{figure*}[ht]
    \centering
    \begin{minipage}{0.32\textwidth}
        \includegraphics[width=\textwidth, clip, trim=75mm 35mm 75mm 35mm]{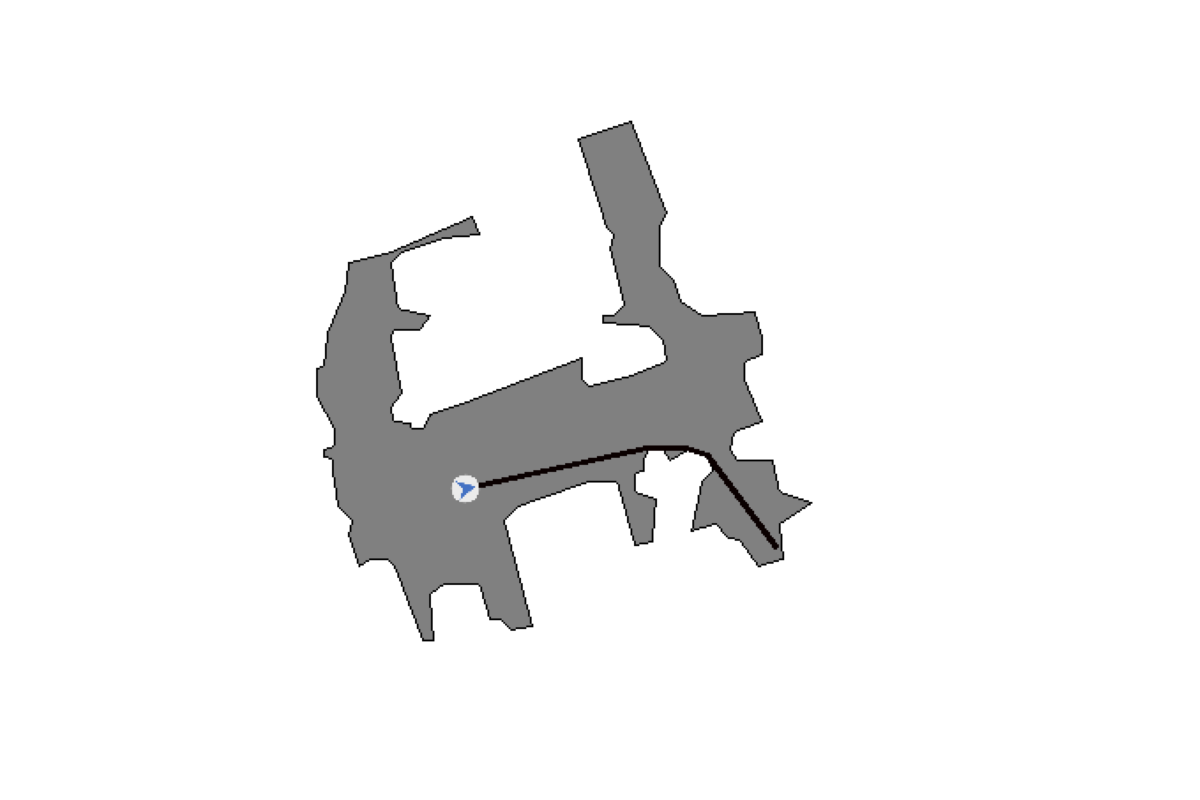}
    \end{minipage}\hfill
    \begin{minipage}{0.32\textwidth}
        \includegraphics[width=\textwidth, clip, trim=75mm 35mm 75mm 35mm]{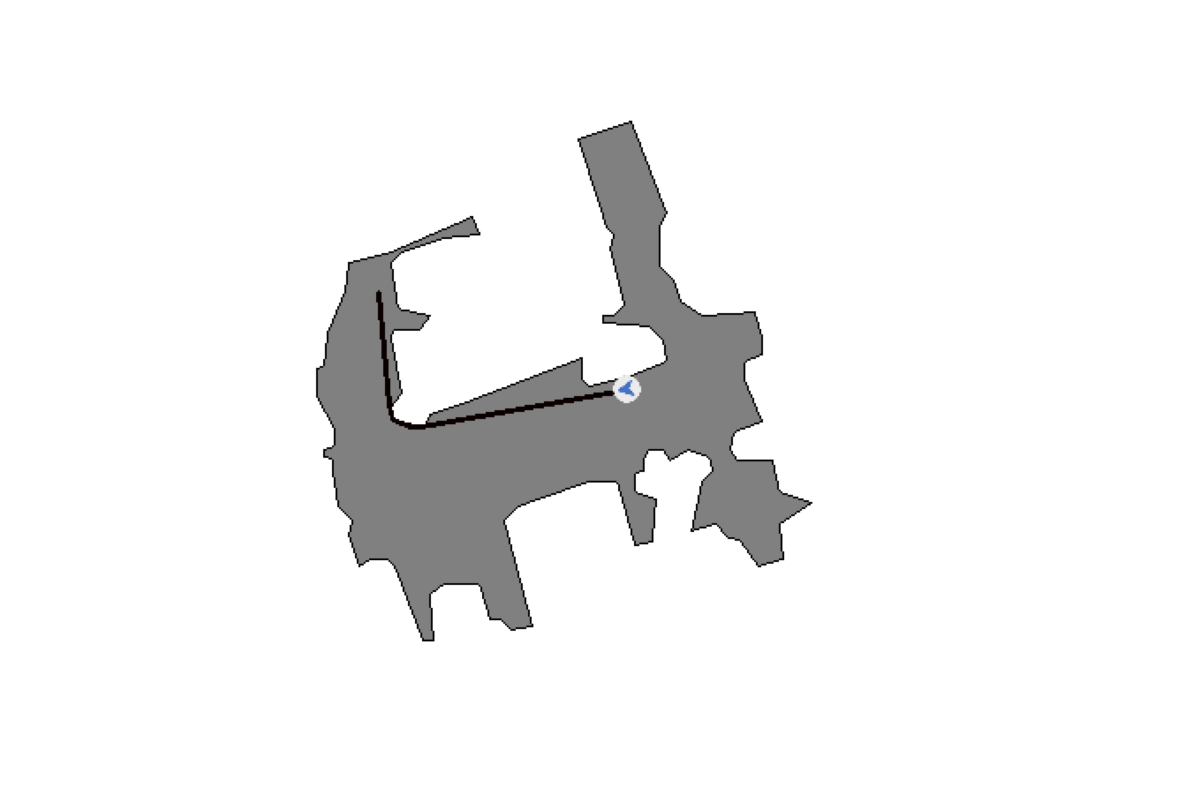}
    \end{minipage}\hfill
    \begin{minipage}{0.32\textwidth}
        \includegraphics[width=\textwidth, clip, trim=75mm 35mm 75mm 35mm]{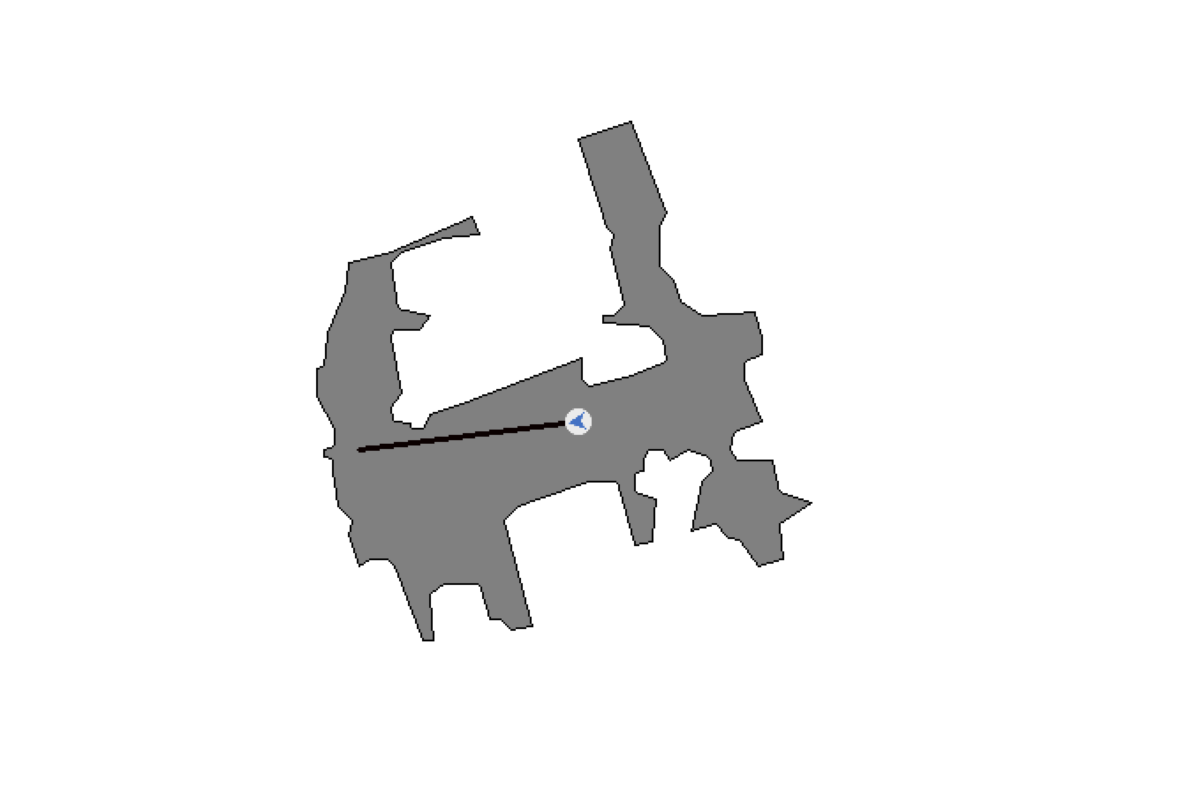}
    \end{minipage}

    \begin{minipage}{0.32\textwidth}
        \includegraphics[width=\textwidth, clip, trim=75mm 35mm 75mm 35mm]{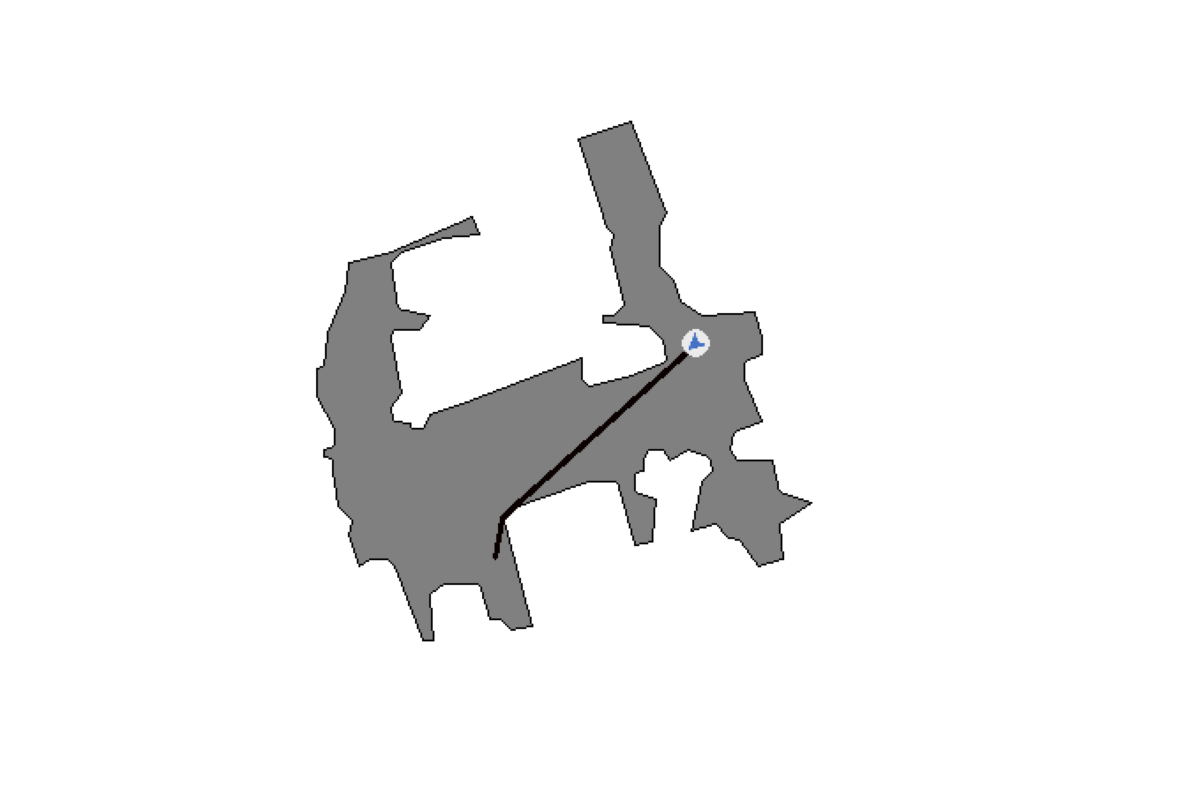}
    \end{minipage}\hfill
    \begin{minipage}{0.32\textwidth}
        \includegraphics[width=\textwidth, clip, trim=75mm 35mm 75mm 35mm]{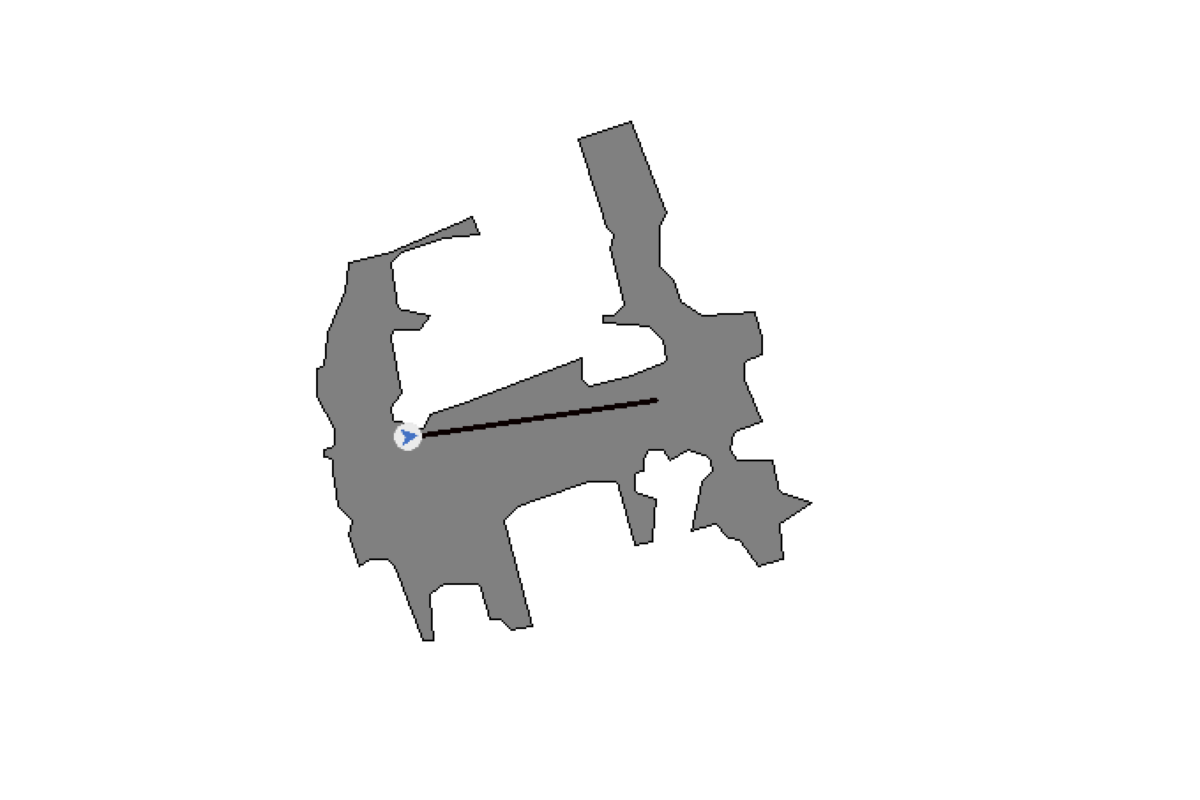}
    \end{minipage}\hfill
    \begin{minipage}{0.32\textwidth}
        \includegraphics[width=\textwidth, clip, trim=75mm 35mm 75mm 35mm]{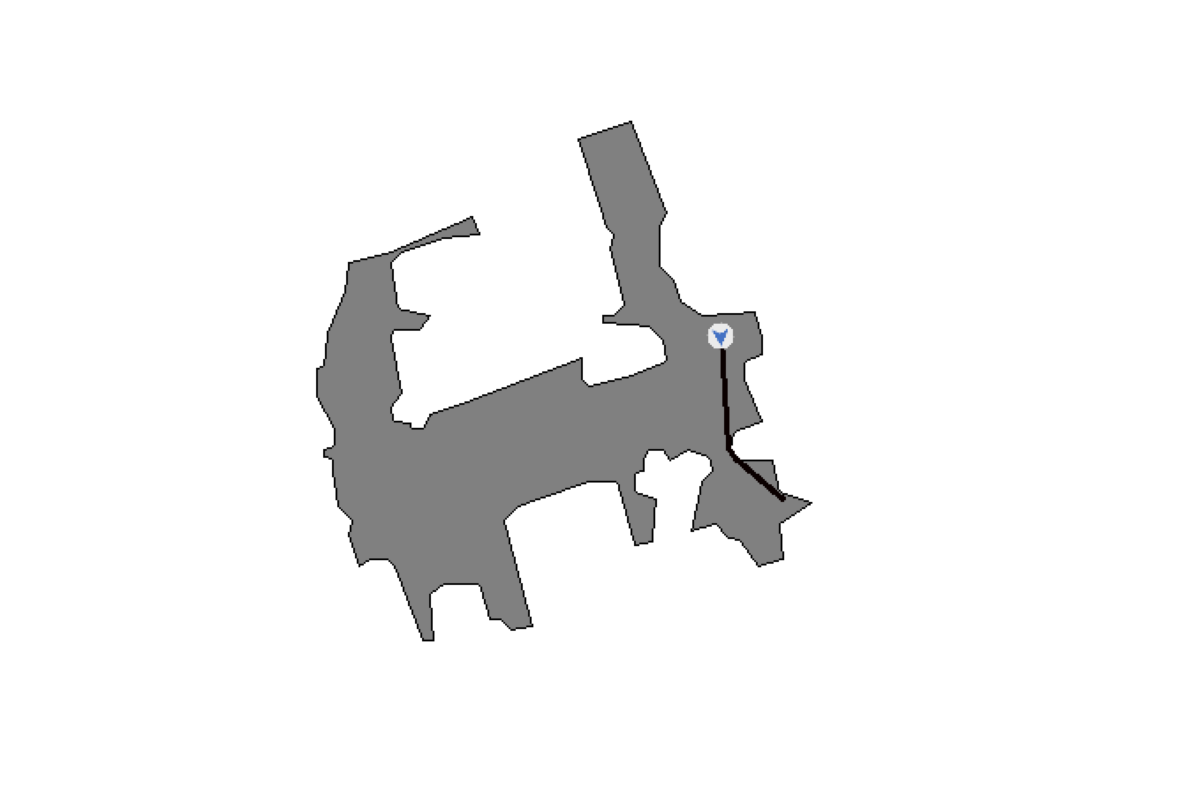}
    \end{minipage}

\caption{Illustrative diverse trajectories generated by the Habitat simulator within a single scene. These demonstrate fundamental navigation actions (e.g., \texttt{``Turn Right''}, \texttt{``Turn Left''}, \texttt{``Turn Back''}), which are foundational for generating route planning QA data via our specialized templates.}\label{fig:habitat_trajectories}
    \label{fig:habitat_sim}
\end{figure*}

\paragraph{Route Plan Data Generation and Template Formatting} 
To generate route planning data, we utilize the Habitat simulator~\cite{habitat} in conjunction with 3D scene data from sources including ScanNet~\cite{dai2017scannet}, ScanNet++~\cite{yeshwanth2023scannet++}, and ARKitScenes~\cite{baruch2021arkitscenes}. Mesh files from these datasets are converted from the \texttt{.ply} format to the \texttt{.glb} format using the assimp library~\cite{assimp2024}. These \texttt{.glb} files then serve as environments within Habitat, where its pathfinder function generates diverse navigation trajectories, as illustrated in Figure~\ref{fig:habitat_trajectories}.

The pathfinder computes a navigable trajectory between two designated points within a scene, outputting a sequence of waypoints. These trajectories are then categorized: those containing turns (defined as a change in direction greater than 30 degrees) are classified as 'with turns,' while others are designated 'without turns.' Trajectories 'with turns' are further labeled as ``Turn Right'' or ``Turn Left,'' depending on the specific angle. For such trajectories, anchor points are defined at the start, the turn itself, and the end. The agent is conceptualized as starting at the initial point, facing the turn point (which serves as a midpoint), and then proceeding to the endpoint. If a turning angle exceeds 45 degrees, an alternative navigation mode is supported where the agent begins at this midpoint, faces the original endpoint, and navigates towards what was the original starting point. Trajectories classified as 'without turns' (i.e., without significant directional changes) are assigned a ``Turn Back'' action. For these ``Turn Back'' paths, anchor points are defined at the start, midpoint, and end of the overall segment; the agent begins at the midpoint, oriented towards the starting point, and then moves to the ending point.

For each anchor point along these generated trajectories, we identify the nearest object using available 3D bounding box annotations from the source datasets. This closest object then provides a semantic description for that anchor point. Once these descriptive anchor points are established for a given path, we formulate the final question-answer templates for our route planning tasks.




    We use \textit{SRC}, \textit{TGT}, and \textit{MID} to represent the object description derived from the annotations. The templates contain two types:
\begin{tcolorbox}[colback=black!5!white,colframe=black!75!black,title=Templates for Route planning]
\textbf{Template 1}\\
"You are a robot beginning at the \textit{SRC} facing the \textit{MID}. You want to navigate to the \textit{TGT}. You will perform the following actions (Note: for each [please fill in], choose either 'turn back,' 'turn left,' or 'turn right.'):  1. Go forward until the \textit{MID}. 2. [please fill in] 3. Go forward until the \textit{TGT}. You have reached the final destination." \\

\textbf{Template 2}\\
"You are a robot beginning at the \textit{MID} facing the \textit{TGT}. You want to navigate to the \textit{SRC}. You will perform the following actions (Note: for each [please fill in], choose either 'turn back,' 'turn left,' or 'turn right.'):  1. [please fill in] 2. Go forward until the \textit{SRC}. You have reached the final destination." \\
\end{tcolorbox}

    For the trajectories with "Turn Right" or "Turn Left", the corresponding description is Template 1 or Template 2. For "Turn Back", the description is Template 2. The descriptions and corresponding actions are paired as question-and-answer (QA) sets and then converted into a multiple-choice format to serve as training data.

\begin{table}[h!]
\centering
\begin{tabular}{|l|r|}
\hline
\textbf{Task Type} & \textbf{Count} \\
\hline
Object Relative Direction & 86,441 \\
Object Absolute Distance & 50,757 \\
Object Relative Distance & 42,025 \\
Object Size Estimate & 12,917 \\
Object Count & 9,357 \\
Route Plan & 4,225 \\
Room Size & 2,057 \\
\hline
\textbf{Total} & \textbf{207,779} \\
\hline
\end{tabular}
\vspace{2mm}
\caption{VSI-Bench Training Data Distribution by Task Type (Total QA pairs: 207,779).}
\label{tab:vsibench_train_distribution_counts}
\end{table}

\subsection{VS\textit{Temporal}I-Bench: New Evaluation Benchmark (138.6K Set)}
\label{sec:appendix_vstibench_design}
\paragraph{Motivation and Design Principles}
Current multimodal models often struggle with spatio-temporal reasoning from monocular video. Even for static scene understanding, accurately interpreting camera motion, along with camera-object and object-object relative movements, remains challenging for LMMs. The VS\textit{Temporal}I-Bench is thus motivated by the critical need to rigorously evaluate and drive progress in LMMs' abilities to comprehend dynamic changes within 3D environments. Our core design principle is to create diverse question-answer pairs that probe the understanding of evolving spatial relationships, such as how camera displacement affects perceived object distances, or how the relative positions of objects change from the camera's perspective over a sequence of frames. The benchmark therefore includes both frame-level tasks requiring precise spatial localization at specific moments and sequence-level tasks demanding an aggregation of information over time to determine motion, direction, or relational changes. This dataset aims to facilitate the examination and advancement of LMMs towards more reliable and human-like visual-spatial-temporal intelligence. Note that this benchmark currently focuses on static 3D scenes, where all depicted motion is from camera movement.

\paragraph{Meta-classes, Task Categories, and Detailed Distribution}
The VSTemporalI-Bench (VSTIbench) is structured into two primary task categories based on the temporal scope of reasoning required: Frame-Level tasks and Sequence-Level tasks. These categories encompass five distinct types of questions designed to probe various aspects of visual-spatial and temporal understanding. The benchmark contains a total of approximately 138.6K question-answer pairs.

The generation of these QA pairs relies on processed metadata derived from raw 3D scene data (point clouds, videos, and sampled frames with depth, instance masks, and camera poses). Specifically, \texttt{scene\_metadata.json} (containing 3D object bounding boxes, instance IDs, scene properties) and \texttt{frame\_metadata.json} (containing per-frame camera poses as 4x4 matrices, camera intrinsics, and 2D object bounding boxes) serve as primary inputs to task-specific generation scripts.

The task categories and their definitions are as follows:

\begin{itemize}
    \item \textbf{Frame-Level Tasks:} These tasks generate questions based on information available within a single frame or the scene's overall point cloud.
    \begin{enumerate}
        \item \textbf{Camera-Object Absolute Distance QA:} Generates questions asking for the approximate Euclidean distance (in meters, numerical answer) between the camera's position in a specific frame and the closest point on the 3D bounding box of a target unique object instance.
        \textit{Generation Logic:} The 3D Euclidean distance is calculated between the camera's world coordinates (from the frame's pose in \texttt{frame\_metadata.json}) and the closest point on the 3D bounding box of the object (derived from \texttt{scene\_metadata.json}).
        \item \textbf{Camera-Object Relative Distance QA:} Generates multiple-choice questions asking which of several candidate object instances is closest to the camera in a specific frame.
        \textit{Generation Logic:} The 3D Euclidean distance is calculated between the camera's position (from the frame's pose in \texttt{frame\_metadata.json}) and the closest point on the 3D bounding box of each candidate object instance (derived from \texttt{scene\_metadata.json}). The candidate object with the minimum distance to the camera is the correct answer.
        \item \textbf{Object-Object Relative Position QA:} Generates multiple-choice questions asking whether a specific target object instance is Near/Far, Left/Right, or Up/Down relative to another unique target object instance, from the camera's perspective in that frame.
        \textit{Generation Logic:} The 3D bounding box vertices of both unique objects (from \texttt{scene\_metadata.json}) are transformed into the camera's coordinate system for the given frame (using the camera pose from \texttt{frame\_metadata.json}). For Near/Far, Z-coordinates are compared. For Left/Right, X-coordinates are compared. For Up/Down, Y-coordinates are compared. Only pairs where one object is entirely nearer/farther, left/right, or up/down than the other (by a defined threshold) are used.
    \end{enumerate}
    \item \textbf{Sequence-Level Tasks (Temporal Tasks):} These tasks generate questions based on information aggregated across a sequence of frames. (The generation logic for these tasks is detailed in the subsequent paragraph).
    \begin{enumerate}
        \item \textbf{Camera Displacement QA:} Generates questions asking for the approximate Euclidean distance (numerical answer) the camera traveled between two specified frames, calculated from the camera's world positions.
        \item \textbf{Camera Movement Direction QA:} Generates multiple-choice questions about the primary direction of the camera's translation during a sequence, relative to its starting orientation. We first compute the overall camera displacement and express it in the starting camera coordinate system, then assign the dominant direction label (e.g., Forward, Backward, Left, or Right). To increase diversity, we instantiate this task as a mixture of 2-choice, 3-choice, and 4-choice questions by sampling different numbers of distractors from the remaining candidate directions. Accordingly, the random baseline is computed by averaging the per-question chance rate over the mixed choice settings, rather than assuming a fixed 4-way setup.
    \end{enumerate}
\end{itemize}





\paragraph{Data Splits (Train/Test)}
The VSTemporalI-Bench comprises a total of 138,610 question-answer pairs, which are divided into training and testing splits as detailed below:

\begin{table}[h!]
    \noindent \textbf{VSTI Train (Total: 132,568 QA pairs)}
    
    \vspace{2pt} 
    
    \centering
    \begin{tabular}{|l|r|}
    \hline
    \textbf{Task Group / Filename} & \textbf{Total Length} \\
    \hline
    \texttt{camera\_displacement} & 32,292 \\
    \texttt{camera\_movement\_direction} & 34,641 \\
    \texttt{camera\_obj\_abs\_dist} & 38,407 \\
    \texttt{camera\_obj\_rel\_dist} & 12,155 \\
    \texttt{obj\_obj\_relative\_pos} & 15,073 \\
    \hline
    \end{tabular}
    \vspace{2mm}
    \caption{VSTemporalI-Bench Training Set Distribution}
\end{table}

\begin{table}[h!]
    \noindent \textbf{VSTI Test (Total: 6,042 QA pairs)}
    
    \vspace{2pt}
    
    \centering
    \begin{tabular}{|l|r|}
    \hline
    \textbf{Task Group / Filename} & \textbf{Total Length} \\
    \hline
    \texttt{camera\_displacement} & 839 \\
    \texttt{camera\_movement\_direction} & 913 \\
    \texttt{camera\_obj\_abs\_dist} & 905 \\
    \texttt{camera\_obj\_rel\_dist} & 1,740 \\
    \texttt{obj\_obj\_relative\_pos} & 1,645 \\
    \hline
    \end{tabular}
    \vspace{2mm}
    \caption{VSTemporalI-Bench Test Set Distribution}
\end{table}

\section{Additional Results and Analysis}
\label{sec:appendix_additional_results}

In this section, we provide additional analysis of VLM-3R from three perspectives. 
First, we examine task-level behaviors on VSI-Bench to understand which capabilities benefit most from reconstructive spatial modeling. 
Second, we compare different pretrained geometry encoders to justify our choice of CUT3R. 
Finally, we evaluate zero-shot generalization on OpenEQA, including comparisons with both the base model and broader multimodal baselines.

\subsection{Task-wise Analysis on VSI-Bench}
\label{sec:appendix_task_analysis}

Our error analysis, together with the quantitative ablation results reported in the main paper, reveals how VLM-3R behaves across different categories of spatial reasoning tasks. 
We observe that the gains from our method are most pronounced on tasks that require explicit geometric perception and metric-aware reasoning.

For example, on \textbf{Absolute Distance}, the full VLM-3R achieves 49.38, substantially outperforming the \texttt{LLaVA-NeXT-Video ft (w/o C\&G Tok.)} baseline, which scores 43.67. 
Removing either camera tokens or geometry tokens also leads to performance drops (48.66 and 49.27, respectively), indicating that both components contribute to accurate absolute-scale distance estimation. 
These results suggest that reconstructive spatial representations are particularly beneficial when the model must reason about metric geometry rather than relying only on appearance cues.

By contrast, on \textbf{Object Counting}, VLM-3R obtains 70.16, which is comparable to both the baseline (70.64) and the variant without geometry tokens (70.30). 
This suggests that, under the current VSI-Bench formulation, object counting depends less on fine-grained 3D geometry and more on 2D visual recognition and cross-frame instance tracking. 
Overall, these observations indicate that VLM-3R mainly improves tasks requiring stronger geometric grounding, while bringing smaller gains on tasks that are already well supported by conventional visual features.

\subsection{Ablation on Pretrained Geometry Encoders}
\label{sec:appendix_encoder_ablation}

To study how the choice of geometric prior affects performance, we compare our default geometry encoder, CUT3R~\cite{wang2025continuous}, with another strong multi-view geometric foundation model, VGGT. 
The results on VSI-Bench are summarized in Table~\ref{tab:encoder_ablation}.

Both geometry encoders improve relative spatial reasoning over the finetuned base model, showing that pretrained geometric priors are broadly beneficial. 
However, the choice of encoder becomes important for tasks involving temporal ordering and absolute metric estimation. 
VGGT adopts a permutation-equivariant design and relies more heavily on relative-scale priors, which makes it less suitable for sequence-aware reasoning and metric-scale judgments. 
As a result, it underperforms on Appearance Order ($38.5 \rightarrow 34.5$) and Room Size ($63.7 \rightarrow 54.0$), despite remaining competitive on some relational tasks.

In contrast, CUT3R naturally preserves temporal structure and provides metric-scale information, which better matches the needs of spatio-temporal reasoning. 
With CUT3R, VLM-3R achieves the best overall performance ($60.9\%$), with particularly clear advantages on Appearance Order ($40.1\%$), Room Size ($67.1\%$), and Route Plan ($45.4\%$). 
These results justify our adoption of CUT3R as the default geometry encoder.

\begin{table}[t]
\centering
\footnotesize
\setlength{\tabcolsep}{4pt}
\renewcommand{\arraystretch}{1.05}
\begin{tabular}{lccc}
\toprule
\textbf{Metric} & \textbf{Base} & \textbf{VGGT} & \textbf{CUT3R} \\
\midrule
Abs. Dist. & 43.6 & \textbf{49.8} & 49.4 \\
Obj. Count      & \textbf{70.6} & 68.8 & 70.2 \\
Obj. Size       & \textbf{70.8} & \textbf{70.8} & 69.2 \\
Room Size       & 63.7 & 54.0 & \textbf{67.1} \\
Rel. Dist.      & 64.9 & 61.5 & \textbf{65.4} \\
Rel. Dir.       & 68.9 & \textbf{80.8} & 80.5 \\
Route Plan      & 40.7 & 44.8 & \textbf{45.4} \\
Appr. Order     & 38.5 & 34.5 & \textbf{40.1} \\
\midrule
Overall         & 57.7 & 58.1 & \textbf{60.9} \\
\bottomrule
\end{tabular}
\vspace{1mm}
\caption{\textbf{Geometry encoder ablation on VSI-Bench.} CUT3R achieves the best overall performance, showing the benefit of metric scale and temporal reasoning.}
\label{tab:encoder_ablation}
\end{table}

\subsection{Generalization to OpenEQA}
\label{sec:appendix_openeqa}

To further assess whether the spatial capabilities learned by VLM-3R transfer beyond VSI-Bench, we evaluate it zero-shot on OpenEQA~\cite{majumdar2024openeqa}, an open-vocabulary embodied question answering benchmark containing 1,600 human-written questions from more than 180 scenes. 
We analyze the results from two perspectives: comparison with the direct base model, and comparison with broader multimodal baselines.

\paragraph{Comparison with the base model.}
A key concern in spatial instruction tuning is whether stronger geometric reasoning comes at the expense of general video understanding. 
To examine this trade-off, we compare VLM-3R directly with its base model, LLaVA-NeXT-Video. 
As shown in Table~\ref{tab:openeqa_base}, VLM-3R improves performance on spatial questions from 49.95\% to 51.60\%, while causing only a modest decrease on non-spatial questions from 67.22\% to 65.54\%. 
Although this leads to a slightly lower overall score, the results confirm that our reconstructive instruction tuning successfully injects stronger spatial awareness into the base model. 
To reduce the loss in general-domain capability, we adopt a mixed-domain training strategy combining \textit{200k VSI-Bench} samples with \textit{30k LLaVA-Video} samples.

\begin{table}[t]
\centering
\footnotesize
\setlength{\tabcolsep}{5pt}
\renewcommand{\arraystretch}{1.05}
\begin{tabular}{lcc}
\toprule
\textbf{Metric} & \textbf{LLaVA-NXT-Video} & \textbf{VLM-3R} \\
\midrule
Spatial      & 49.95 & \textbf{51.60} \\
Non-Spatial  & \textbf{67.22} & 65.54 \\
Overall      & \textbf{62.4} & 61.7 \\
\bottomrule
\end{tabular}
\caption{\textbf{Base-model comparison on OpenEQA.} VLM-3R improves spatial performance over its base model, with a small trade-off on non-spatial questions.}
\vspace{-2mm}
\label{tab:openeqa_base}
\end{table}

\begin{table}[t]
\centering
\footnotesize
\setlength{\tabcolsep}{6pt}
\renewcommand{\arraystretch}{1.05}
\begin{tabular}{@{}r l c | r l c@{}}
\toprule
 & \textbf{Method} & \textbf{Perf. (\%)} & & \textbf{Method} & \textbf{Perf. (\%)} \\
\midrule
1 & Human              & 86.8 & 5 & Gemini-Pro & 44.9 \\
2 & \textbf{VLM-3R}    & \textbf{61.7} & 6 & Claude 3   & 36.3 \\
3 & GPT-4V (50 frames) & 55.3 & 7 & GPT-4       & 33.5 \\
4 & GPT-4V (15 frames) & 54.6 & 8 & LLaMA2      & 28.3 \\
\bottomrule
\end{tabular}
\caption{\textbf{Zero-shot performance against general baselines on OpenEQA.} VLM-3R remains highly competitive among strong multimodal baselines.}
\vspace{-5mm}
\label{tab:openeqa_sota}
\end{table}

\paragraph{Comparison with broader multimodal baselines.}
Beyond the base-model comparison, VLM-3R also remains competitive against strong general-purpose multimodal models. 
As shown in Table~\ref{tab:openeqa_sota}, VLM-3R achieves 61.7\% overall zero-shot accuracy, outperforming GPT-4V, Gemini-Pro, Claude 3, GPT-4, and LLaMA2. 
These results suggest that the proposed geometry-aware fusion and mixed-domain training strategy improve spatial understanding without sacrificing competitiveness on a broader embodied benchmark.